\documentclass[10pt,journal,compsoc]{IEEEtran}

\ifCLASSOPTIONcompsoc
  \usepackage[nocompress]{cite}
\else
  \usepackage{cite}
\fi
\ifCLASSINFOpdf
\else
\fi

\usepackage{multirow}
\usepackage{amsthm,amssymb,amsmath,graphicx}

\usepackage{color}
\usepackage{amsmath}
\usepackage{xcolor}
\usepackage[shortlabels]{enumitem}

\newcommand{\specialitem}[3][white]{%
  \item[%
    \colorbox{#2}{\textcolor{#1}{\makebox(14,14){#3}}}%
  ]
}

\definecolor{editorOrange}{cmyk}{0, 0.8, 1, 0}
\definecolor{editorBlue}{cmyk}{1, 0.6, 0, 0}
\definecolor{editorGreen}{cmyk}{0.66, 0, 0.87, 0}
\definecolor{editorPink}{cmyk}{0, 1, 0, 0}
\usepackage[caption = false]{subfig}

\begin{document}
%
\title{MedicalRec: Medical recommender system for image classification without retraining}
%
%
%
%
\author{Roghayeh Taghavi, Aysa Hasanazde Bashkandi, Amir Ali Bengari, Mohammad Amin Raji, Mohammad Salahi Ardekani, Parisa Mardukhian, Parvaneh Rezaei, Ramin Mousa}

\markboth{Journal of \LaTeX\ Class Files,~Vol.~14, No.~8, August~2015}%
{Shell \MakeLowercase{\textit{et al.}}: Bare Advanced Demo of IEEEtran.cls for IEEE Computer Society Journals}
%



\IEEEtitleabstractindextext{%
\begin{abstract}
The emergence of machine learning and deep learning has revolutionized the efficiency of diagnostic, therapeutic, and administrative systems in healthcare. However, this rapid adoption has come at the cost of requiring significant computing power and energy consumption, as well as e-waste disposal and carbon emissions. One of the challenges of these models is choosing the right model for classification tasks. To this end, researchers attempt to identify the optimal model using their data through trial and error, which involves energy consumption and waste. The goal of this study is to develop a model-based recommender system for medical image classification. For this purpose, a data set was collected from 3,000 articles in the field of medical image classification. This dataset, publicly available under the name MedicalRec-Bench, contains over 5,000 records of models tested in various tasks, including Skin Cancer Classification, Tumour Classification, Wound Classification, Breast Cancer, and MRI classification. The dataset was evaluated in four different modes, depending on the number of features: MedicalRec I (5 features), MedicalRec II (9 features), MedicalRec III (11 features), and MedicalRec IV (18 features). Collecting all values for the features is challenging due to non-reporting by the authors; hence, the dataset contains significant amounts of missing values. The Medical Recommender System (MedicalRec) is a transformer-based model used for item recommendations in this study. This model achieved remarkable results in the evaluation on the dataset and in the evaluation with 12 base models. This model achieved a maximum HitRate@100 of 75.5\%. The dataset and implementations are available through the GitHub link: \textcolor[rgb]{0.94,0.00,0.94}{https://github.com/Ramin1Mousa/MedicalRec}.
\end{abstract}

}

\maketitle

\IEEEdisplaynontitleabstractindextext

%
\IEEEpeerreviewmaketitle

\section{Introduction}
\label{sec:introduction}
\IEEEPARstart{M}{edical} image classification plays a crucial role in the analysis and diagnosis of diseases, as well as in clinical care and treatment. Traditional machine learning and deep learning have achieved extraordinary results in medical image classification over the past few decades as learning tools. These models have been used in the classification of computed tomography (CT), magnetic resonance imaging (MRI), positron emission tomography (PET), mammography, ultrasound, X-ray, and other modalities. In machine learning models, the goal is to map input to output using a single mapping layer, which enables fast learning. In contrast, deep learning models attempt to map input to output through multiple layers, allowing for slower learning but achieving higher accuracy. The increasing energy consumption and carbon footprint of deep learning (DL) have become a concern due to the increasing computational requirements \cite{M1}. 

The IPCC-2021 Assessment \cite{inT1} indicates that global warming will increase by 1.5 to 2°C during the 21st century, which will exceed the allowable limit. The solution lies in reducing carbon dioxide (CO2) and other greenhouse gases in the coming decades. Deep learning models, as a relatively small but rapidly growing field in medical image analysis (MIA), play a significant role in carbon production. The availability of big data and massive computational resources has accelerated the advancement of DL methods. According to estimates \cite{inT2}\cite{inT3}, the computation required for machine learning methods has doubled every 3.4 to 6 months since 2010.

The challenge in medical image classification is to select the appropriate model for classification. New models are being introduced daily, incorporating the latest methods. In many studies presented in the literature, providers evaluate dozens of models using their data to identify the most effective one. This challenge, in addition to the time spent, leads to carbon production and energy waste. In this study, the goal is to develop a system that recommends the best models for a dataset to the user from among the models available in the literature. The lack of a suitable protocol for providing a classification model is cumbersome. For this purpose, MedicalRec-bench is proposed in this study. MedicalRec-bench is a dataset containing features extracted from 3000 articles in medical image classification. This dataset includes title, keywords, abstract, statistical features of the dataset, evaluation metrics, and the best model presented in each study. For this purpose, 19 features were extracted from each study. This dataset was made publicly available in four versions: MedicalRec I, MedicalRec II, MedicalRec III, and MedicalRec IV.

Selecting the right model for each dataset is actually an item representation problem that falls under the category of recommender systems. Recommender systems play a vital role in reducing redundant information to enrich the online experience \cite{Intro1}\cite{Intro2}. They provide personalized recommendations for relevant candidate items across various application domains, including entertainment \cite{Intro3}, e-commerce \cite{Intro4}, and job matching \cite{Intro2}. The primary concept of recommender systems is to utilize the interactions between users and items, along with their associated contextual information, to predict the matching score between users and items \cite{Intro5}. Due to their remarkable ability to learn representations in various fields, deep neural networks (DNNs) have been widely used for advanced recommender systems.

In this study, MedicalRec was introduced to suggest models based on inputs. MedicalRec is essentially a Transformer-based model that utilizes the Softmax function in its last layer to calculate the score of each item (model). In this model, Bert was used to embed the input data. The results obtained show the superiority of this model over twelve baseline models.

\section{Related work}

The emergence of artificial intelligence (AI) has transformed healthcare by improving diagnostic, treatment, and administrative efficiency. This tool is currently being utilized by radiologists, cardiologists, oncologists, and other medical professionals to support their decision-making and reduce errors \cite{[1]}.

However, this rapid adoption come with an impact on the environment: requiring substantial computational power and energy consumption, disposing of electronic waste, and carbon emission \cite{[2]}.
Carbon emissions are becoming a global concern as more deep learning (DL) models are being developed and deployed. Training a state-of-the-art (SOTA) model demands a significant amount of energy, which contributes to a considerable amount of greenhouse gas (GHG) emissions \cite{[3]}.

\subsection{Carbon Footprint of Training Deep Learning Models}
To reduce emissions, machine learning (ML) practitioners should utilize “Green AI” in their work. Green AI refers to research in the AI field that considers computational cost, resource usage, and carbon emission while achieving excellent results. In contrast, “Red AI” refers to research that use substantial computational power and energy to improve accuracy and SOTA results \cite{[4]}.
Although “Red AI” work is valuable and ground-breaking, the total carbon footprint of training different models has quadrupled since 2016, and the primary sources of these emissions are the carbon intensity of the electricity infrastructure and the training time \cite{[5]}.
Considerable training duration to achieve high accuracy and SOTA results demand considerable resources, leading to significant environmental costs. Strubell et al. reported that training a large DL model with neural architecture search (NAS) emits over 626,000 lbs. CO2 equivalent, approximately five times the lifetime emissions of a typical car [3]. Moreover, development of convolutional neural networks (CNNs), extensively used in medical image analysis, contributes substantially to carbon emissions. This is compounded with large medical imaging datasets (e.g., chest x-rays, computed tomography, magnetic resonance imaging), and the frequent retraining for different tasks. Together, the energy demand of CNN-based models could easily escalate and leave a negative impact on global carbon emissions, without drastic gains in performance \cite{[6]}\cite{[7]}.

Several methods and strategies that can help reduce carbon footprint are: 1) quantification of carbon footprint using frameworks such as Carbontracker and Eco2AI during training phase, 2) energy-efficient algorithms and architectures using federated learning (FL), tensor networks (TN), and transfer learning, and 3) green-energy infrastructure such as carbon-neutral energy, carbon-credit, and carbon offsetting. The lack of standardized metrics to measure carbon footprint and the absence of transparent lifecycle assessments of computing systems are challenges that the tech industry and ML practitioners should focus on \cite{[8]}.

Despite substantial advances in reducing training-phase emissions and improving inference efficiency, the environmental impact of frequent retraining of models for new DL tasks – especially in domains like medical imaging – remains underexplored. This gap requires solutions to maximize the utility of existing pre-trained models and diminish the carbon costs of retraining and operational decisions.

In the current section, we review several related works in following the methodology of 'green AI' and 'sustainable AI', including carbon tracking tools, mitigation strategies, and other works.

Efforts to quantify the environmental impact of DL training have led researchers to the development of several open-source carbon tracking tools. CarbonTracker, a Python package, is designed to monitor and predict the energy consumption and carbon emissions during training in real-time. It considers the number of training epochs, hardware configuration, and geographical location of computing. The tool has been validated on convolutional neural network (CNN) architectures applied to medical image segmentation tasks \cite{[9]}.

Similarly, Eco2AI offers a Python library for real-time tracking of energy usage and carbon emissions, contributes to “Sustainable AI” by providing carbon intensity coefficients for 365 geographic regions. In their evaluation with Malevich and Kandinsky text-to-image models with different sampling strategies, the authors showed that a 3-bit Gaussian Error Linear Unit (GELU) configuration on Malevich, reduced the carbon emissions by 17\% compared to the full-precision version \cite{[10]}.
Other frameworks like the Experiment impact tracker (EIT) and CodeCarbon have expanded real-time monitoring to include transparency features like leaderboards and regional performance comparisons. EIT encourages more ML researchers toward the adoption of sustainable practices, such as energy-efficient algorithm design, regional optimization based on renewable energies, and promotes transparency in reporting \cite{[7]}.
Recent comparative studies have highlighted the differences among these tracking frameworks. One study by Verma et al. compared Temporal Fusion Transformer (TFT), Attention-based Long Short-Term Memory (Att-LSTM), and LSTM models using datasets from World Bank Indicators (WBI) and Our World in Data (OWID). While TFT showed the lowest carbon emissions, LSTM had the shortest training time among the three models. Interestingly, CarbonTracker reported lower power consumption for the LSTM model, contrary to the other tools, demonstrating the nuances among carbon tracking tools and importance of tool selection \cite{[11]}.
A review by Bannour et al. further revealed inconsistencies among tracking frameworks applied to Natural Language Processing (NLP) tasks, specifically Named Entity Recognition (NER). These discrepancies were linked to variations in carbon intensity calculations, hardware availability, and tool precision. They recommend using online tracking tools for accessibility and ease of integration \cite{[12]}.

Lacoste et al. proposed several practices to reduce the emissions of ML experiments, including efficient hyperparameter searches, hardware selection, data center location, and cloud service providers. While cloud providers recently adopted carbon-neutral strategies, utilizing monitoring frameworks further helps minimize the carbon footprint. The Software Carbon Intensity (SCI) metric proposed by Dodge et al. provides ML practitioners a standardized method to evaluate the environmental impact of both training and inference on cloud servers. However, the embodied emissions, such as hardware production and disposal, remain unaccounted for \cite{[13]}\cite{[14]}.
\subsubsection{Mitigation strategies}
Several studies have explored practical approaches to reduce the carbon footprint of DL models. Knowledge Distillation (KD), is a strategy to train student models using larger teacher models. Rafat et al. proposed a stochastic framework that preserved performance with different image classification and object detection tasks using three models – ResNet18, MobilNetV2, and ShuffleNetV2. Their method helped reduce energy consumption and carbon emissions while keeping performance similar to the original setup \cite{[15]}.

A different direction is to decentralize DL models using Federated Learning (FL) which comes with its challenges. Depending on the configuration, FL can have significantly higher emissions compared to centralized learning – particularly hardware inefficiency, poor aggregation strategy, high overhead communication, and imbalance data distribution. These results emphasize the need for thoughtful system design to achieve comparable efficiency and sustainability goals \cite{[16]}.
In a comprehensive review of over 200 publications (from 2014 to 2024) by Hasan et al., they provide practical tools and theoretical methods to reduce carbon emissions. They propose solutions such as model compression, green computing, smart grids with ML, renewable energy integration, and carbon-aware design. The review highlights the need for standardized metrics, clearer sustainability of large-scale green models, and the compromise between security and emissions as challenges on the road to a greener AI community \cite{[8]}.
\subsubsection{Advancements and Gap}
Carbon emissions from ML development can be categorized into “Operational” and “Lifecycle” emissions. While lifecycle emission is embedded in the manufacturing, transfer, and disposal of all components, operational emissions could be reduced by practices such as using efficient models and components and cloud computing. Furthermore, measurements of energy consumption and emissions provided by studies in the field, while valuable, underestimate future advances in hardware, model algorithms and architectures. The biggest challenge in this field is not the operational emissions of ML, but the lifecycle cost of manufacturing of all equipment, which needs more transparency and accountability from hardware manufacturers. However, ML practitioners should not overlook their role in reducing emissions from their experiments, contributing to “Green AI” practices \cite{[17]}.
While advancements in ML research reduce energy consumption and emissions, these improvements often result in a rebound effect known as the “Jevons Paradox”. As efficiency in the technology increases, it becomes more attractive and more popular, resulting in increased resource consumption and emissions. This effect, combined with the growing demand for models with billions of parameters and ever-higher accuracy, can offset or even exceed the environmental gains achieved through efficiency measures \cite{[18]}.

\subsection{Foundational and Survey Studies on Recommender Systems}
In general, a recommender system is an intelligent information filtering framework designed to estimate or predict a user’s potential interest or rating for specific items. These systems play a crucial role in assisting users to discover relevant content across a wide range of domains, including music and video streaming services, online shopping platforms, and personalized content delivery. Beyond entertainment and e-commerce, recommender systems are increasingly being applied in research discovery, expert collaboration environments, and financial decision-making support, highlighting their versatility and growing importance in modern information systems. Transformer-based recommender system research is still a relatively new and growing field, with a major focus on item and event recommendation. China is at the forefront of this field with its massive data and platform-independent ecosystem, and there is still plenty of opportunity to develop more accurate and innovative methods \cite{RS1}.

Diffusion Models are among the practical models that have been used in the implementation of recommender systems. These models operate by progressively adding noise to the input data during a forward process and then training a neural network to recover the original data from these noisy samples. During the inference phase, the trained model iteratively predicts and removes portions of noise from an initial random input until a clean output is reconstructed. Although diffusion models were initially popularized for their impressive image generation capabilities, their strength lies in accurately modeling complex data distributions. By learning the underlying structure of the data, these models can effectively generate new samples that follow the same distribution \cite{RS2}.

Diffusion models have been increasingly applied to various recommendation tasks. In collaborative filtering, they are utilized to model both implicit and explicit user feedback, often incorporating user–item graphs to better capture interaction structures. Regarding sequential recommendation, diffusion processes can treat user behavior sequences as either the diffusion target or guidance, enabling effective modeling of temporal dependencies in point-of-interest (POI) or session-based scenarios. Within multi-domain recommendation, diffusion models facilitate cross-domain learning by integrating multi-modal attributes, such as visual or textual features, and even enabling text-to-recommendation or image-generation tasks. Furthermore, in responsible recommendation, diffusion-based frameworks have been explored to improve fairness, accountability, transparency, and robustness to out-of-distribution data \cite{RS2}.

Large Language Models (LLMs) have paved the way for a new generation of generative recommender systems (Gen-RecSys) that are able to generate textual or multi-modal responses rather than presenting limited lists of items (Top-K). These responses can incorporate user commands and context, providing a more natural and personalized experience for the user. Despite these advantages, generative systems have also created new challenges, including content hallucinations, private information leakage, and new forms of bias and persuasion. Therefore, the evaluation of these models can no longer rely solely on classic indicators such as ranking accuracy (top-k metrics), but requires a more comprehensive approach that also includes aspects of factuality, alignment, fairness, and safety \cite{RS3}.

\subsubsection{Transformer-based and Sequential Recommendation Models}

In most real-world recommender system scenarios, user interaction data exhibits two key characteristics: sequential and multi-behavioral patterns. A simultaneous understanding and modeling of these properties are crucial for improving the accuracy and personalization of recommendations. Accordingly, modern recommender models leverage advanced architectures such as Transformers and attention mechanisms to more effectively capture the dynamics of user behaviors and the diversity of their interactions.
BERT4Rec is a sequential recommender model based on the Transformer architecture, which leverages bidirectional self-attention to better capture dependencies among items. Unlike RNN-based approaches that process user interactions in a single direction, BERT4Rec learns contextual relationships in both directions through the Cloze (masked item prediction) task. The model is composed of multiple stacked Transformer layers, where each layer iteratively refines the vector representation of every item by exchanging information across all sequence positions. Each layer includes two main components: Multi-Head Self-Attention, which learns inter-item relationships along the sequence, and a Position-wise Feed-Forward Network, which enhances feature representations at each position. This architecture has demonstrated significant improvements in recommendation accuracy across multiple real-world datasets\cite{RS4}.

BERT\_Music is a music recommender system built upon the Transformer architecture and the BERT model, which employs bidirectional self-attention to learn contextual and sequential dependencies among songs listened to by the user. As users’ musical preferences evolve over time, accurate prediction becomes challenging; therefore, this model is specifically designed to capture and represent dynamic user preferences. The Echo Nest and ListenBrainz datasets were used for training and evaluation, containing users’ listening histories and song play counts. Experimental results demonstrate that BERT\_Music outperforms previous approaches in sequence prediction and playlist continuation tasks \cite{RS5}.

In terms of multi-behavioral patterns, the MB-STR (Multi-Behavior Sequential Transformer Recommender) framework is designed to overcome the limitations of previous models in modeling multi-behavior dependencies, identifying diverse sequential patterns, and reducing data fragmentation. The model consists of three key components: the Multi-Behavior Transformer (MB-Trans) layer to model the dependencies and meaning of each behavior, the Multi-Behavior Sequential Pattern Generator (MB-SPG) to encode diverse patterns, and the Behavior-Aware Prediction (BA-Pred) module to effectively exploit multi-behavior data. Experiments on several real datasets show that MB-STR outperforms previous methods, and its modules play a key role in improving the accuracy of recommendations \cite{RS6}.

Yuan, Enming, et al. ”Multi-behavior sequential transformer rec ommender.” Proceedings of the 45th international ACM SIGIR conference on research and development in information retrieval. 2022.

The GLINT-RU model is a lightweight and efficient sequential recommender system designed to reduce computational costs and accelerate prediction. It employs a Dense Selective GRU that adaptively models temporal dependencies and fine-grained positional information through Selective Gates. A Parallel Mixing Block injects precise positional features into user–item interactions, enhancing the quality of latent representations. The model architecture consists of an item embedding layer, Dense Selective GRU, and a selective MLP block, collectively providing a rich representation of user behavior. Experiments on three real-world datasets demonstrate that GLINT-RU achieves both higher prediction accuracy and faster training and inference compared to traditional and other efficient SRS models. These results establish GLINT-RU as an innovative, stable, and effective framework for sequential recommendation tasks \cite{RS7}.

Zhang, Sheng, et al. ”Glint-ru: Gated lightweight intelligent recur rent units for sequential recommender systems.” Proceedings of the 31st ACM SIGKDD Conference on Knowledge Discovery and Data Mining V. 1. 2025. 

To improve recommendation accuracy, combining diverse information sources is a logical approach; however, most studies rely solely on the utility matrix and do not integrate textual sources with it. To address this limitation, we propose the Multiview Transformer model, which efficiently integrates textual information with the utility matrix. This model employs the user–item utility matrix along with textual data, including user reviews, item descriptions, and categories, in a multiview manner. Features from the utility matrix are extracted via matrix decomposition and ALS, reviews and textual descriptions are converted into feature vectors using BERT, and item categories are modeled with TF-IDF. The resulting features are then transformed into a unified representation and fed into a Transformer model to predict ratings or generate recommendations. Experiments on the Amazon and Movielens datasets demonstrate that the proposed model achieves higher prediction accuracy and significant reductions in MAE and RMSE compared to baseline and graph-based models. Consequently, Multiview Transformer effectively integrates utility matrix and textual data to model complex user–item dependencies and improve recommendation quality \cite{RS8}.

\subsubsection{Federated and Privacy-Preserving Recommendation Frameworks}
Federated recommender systems enable training of models in a distributed environment, where user data is retained locally on devices and does not need to be transmitted to a central server, effectively safeguarding user privacy. In these systems, there are typically two main roles: clients, which are local user devices utilizing their data for model updates, and a central server, which aggregates the locally updated parameters from multiple clients to update the global model.

The DGFedRS model is a federated sequential recommender system that leverages a pre-trained diffusion model to expand user–item interaction sequences and enrich sparse data. It employs four key strategies: diffusion augmentation to extend interactions, guided denoising to generate high-quality samples, noise control to preserve user-specific preferences, and a stepwise scheduling mechanism to feed generated data into the recommender model effectively. Historical user interactions are compressed into low-dimensional vectors using k-means clustering and variational encoders to reduce computational costs. The federated learning process involves local training on user devices and aggregation of model updates on a central server, ensuring data privacy. Experiments on three real-world datasets demonstrate that DGFedRS achieves superior performance while generating personalized user interactions. Future work focuses on model compression and scenario-adaptive architectures to enhance scalability and adaptability across diverse recommendation contexts \cite{RS9}.
Another study by Reddy et al. presents an innovative framework for recommender systems that combines federated learning (FL) with transformer models such as BERT and Behavior Sequence Transformer (BST) to enhance prediction performance while preserving user privacy. In this approach, models are trained locally on user data, and only updated parameters are transmitted to the central server, without sharing raw user data. The Adaptive Federated Optimization (FedAvgOpt) algorithm optimizes the global training process by leveraging various optimizers, including Yogi, AdaGrad, and Adam, enabling effective learning in heterogeneous and distributed environments. Experiments on the Amazon Customer Review and Movielens-1M datasets demonstrate that the federated BERT model achieves outstanding performance with accuracies of 87\% and 76\%, while the BST model attains an MAE of 0.8. This research highlights that integrating federated learning with transformers can significantly improve recommendation accuracy without compromising user privacy. Future work will focus on practical deployment of federated models in real-time systems, addressing data heterogeneity and communication overhead to achieve secure, personalized, and efficient recommender systems \cite{RS10}.

Recommender systems are widely used across industries to enhance prediction accuracy and user experience by leveraging user–item interactions, yet the increasing number of items and user data raises significant privacy concerns. To address this challenge, we propose LDPMF, a local differential privacy (LDP)-based recommender system that protects both user ratings and evaluated items. LDPMF employs a stochastic process combined with dimensionality reduction to reduce perturbation errors and stabilize noisy gradients during iterative updates. Matrix factorization generates user and item latent vectors, while the LDP mechanism ensures privacy protection throughout all gradient descent iterations. Experimental results on the MovieLens and LibimSeTi datasets demonstrate that LDPMF achieves higher prediction accuracy, effectively preserves user privacy, and reduces communication costs. Overall, LDPMF provides a robust, privacy-preserving framework that enhances both the effectiveness and efficiency of recommender systems \cite{RS11}.
\subsubsection{LLM-driven, Conversational, and Generative Recommender Systems}
Conversational Recommender Systems (CRSs) enable interactive, dialogue-based recommendations, yet most existing approaches overlook the potential sequential dependencies between items and entities discussed in a conversation. With the advancement of large language models and chatbots, CRSs are increasingly capable of delivering personalized recommendations through natural, context-aware dialogues, highlighting the need for models that capture sequential and generative aspects of user interactions. In a study by Zou et al., a transformer-based TSCR method is introduced to model these sequential dependencies and improve the performance of the recommender system. Conversations are represented as a sequence of items and entities, and user preferences are extracted from these sequences. The Cloze task is used to train the model, and the hidden items in the sequence are predicted to use the two-way information from the conversation sequence. A deep two-way self-attention architecture is used, and the related entities are extracted from the knowledge base, without using the knowledge base structure for reasoning, which makes the method simple and straightforward. The experimental results on the ReDial dataset have shown that the TSCR model, despite its simplicity, is very effective and is recognized as a strong basis for future research in the field of conversational recommender systems \cite{RS12}.

In another study, a framework called CSHI (Controllable, Scalable, and Human-Involved) was introduced, which employs a Plugin Manager to control the behavior of user simulators at different stages. In this framework, the simulation of user behavior and interaction is designed to be customizable and scalable, aiming to create a more natural and human-like user experience. The simulation process consists of three main stages: user profile initialization, preference initialization, and message management, which are implemented through either independent or cooperative plugins. To prevent data leakage and produce more realistic user feedback, mechanisms such as user memory simulation, segregation of known and unknown preferences, and anonymization of sensitive information are applied. The response generation plugins interpret the intent of the recommender system and generate appropriate feedback, including answering questions, accepting or rejecting recommendations, and engaging in natural dialogues. The results of experimental evaluations and case studies demonstrated that the CSHI framework can effectively adapt to various conversational recommendation scenarios and generate user feedback closely resembling real user behavior, thereby facilitating the evaluation of conversational recommender systems and the creation of high-quality conversational datasets \cite{RS13},
Another study adopts a critical perspective on previous studies of fairness evaluation in recommender systems based on large language models (RecLLMs). Earlier approaches have typically assessed fairness by comparing recommendation lists generated with and without users’ sensitive attributes, without adequately distinguishing between genuine personalization and algorithmic bias. To address these limitations, we introduce CFaiRLLM, a framework that not only measures “alignment with users’ true preferences” but also extends the notion of “intersectional fairness” by accounting for the overlap of multiple sensitive attributes. The framework further incorporates diverse user profile sampling strategies—random, top-rated, and recency-based—to analyze how different modes of profile representation influence fairness in RecLLMs. Experimental results on the MovieLens and LastFM datasets demonstrate that true preference alignment provides a more accurate and equitable assessment than similarity-based measures. Moreover, the findings reveal that intersectional attributes amplify fairness disparities, particularly in less structured domains such as music, underscoring the importance of integrating true preference alignment into fairness evaluation for RecLLMs \cite{RS14}.

\subsubsection{Emerging Learning Paradigms in Recommender Systems: Reinforcement Learning and Causal Approaches}
As key enablers of personalized services, recommender systems have traditionally operated based on data correlations. However, relying solely on correlations without accounting for the underlying causal relationships can lead to issues such as bias, lack of transparency, reduced trustworthiness, and fairness concerns. To address these limitations, recent research has introduced the application of causal inference to enhance both the performance and fairness of recommender systems. Causal inference, grounded in the frameworks of potential outcomes and structural causal models (SCMs), enables a deeper understanding of the hidden mechanisms that drive decision-making processes. The integration of causal methods into recommender systems not only improves predictive accuracy but also enhances the transparency, explainability, and fairness of recommendations. Methods such as data reweighting, counterfactual reasoning, structural equation modeling, and causal graph analysis are among the principal approaches used to mitigate bias and ensure fairness. Finally, emerging studies suggest that combining reinforcement learning with causal inference can pave the way for a new generation of intelligent, fair, and explainable recommender systems \cite{RS15}.

Recent advances in recommender systems have increasingly adopted reinforcement learning and causal inference as emerging paradigms to achieve adaptive, explainable, and user-centric recommendations. In a study by Tzeng et al., a personalized exercise recommendation system for Massive Open Online Courses (MOOCs) was developed using an Actor–Critic reinforcement learning framework. The system analyzes students’ online learning trajectories and provides suitable exercises via a LINE bot interface. By integrating objectives related to review, difficulty, and learning progress, and by designing a reward function within the reinforcement learning paradigm, the system effectively increased students’ motivation to participate and complete exercises—the completion rate among system users reached 89.97\%, compared to 47.23\% in the control group. The model recommends exercises with appropriate difficulty levels and conceptual coverage for each student and is easily accessible through LINE. Empirical evidence, including midterm grades and 227 survey responses, indicated improved learning effectiveness and high student satisfaction (approximately 90\%). The system estimates value functions and optimal exercise selection policies using deep neural networks and Markov decision processes. Overall, the results demonstrate that the proposed framework successfully promotes desirable learning behaviors and enhances the efficiency of personalized learning. Future work aims to integrate natural language processing techniques to enable richer interactions with LINE users \cite{RS16}.

In another study, human-centered decision-making supported by artificial intelligence in the form of recommender systems is examined. The study aims to analyze the role of human judgment and shared situational awareness between humans and automated agents (AI) in joint decision-making processes. To this end, a framework was developed to enhance human situational awareness and improve judgment by providing contextual information utilized by the intelligent system. A two-phase experiment was conducted involving human participants and an automated recommender system in a high-stakes decision-making scenario, where the amount of available information, evaluation method, and system reliability were systematically varied. The findings indicate that increasing the transparency of the information used by the system enhances shared situational awareness, improves team performance, and calibrates the level of trust in the AI agent. Moreover, active human involvement in evaluating the system’s recommendations reduced overreliance on AI and improved understanding of its limitations. This approach provides an effective means of enhancing model transparency and fostering human–AI collaboration in high-risk recommender systems \cite{RS17}.

In modern recommender systems, user and item features are typically represented as embeddings. However, these embeddings often entangle intrinsic features (e.g., the user’s actual interest in an item) with popularity factors (e.g., preference for popular items), which can degrade model performance when popularity distributions change over time. Yoo et al. introduce a novel framework called TPAB (Temporal Popularity Distribution Shift Generalizable Recommender System) to address temporal changes in popularity. TPAB leverages three innovations—time-aware embeddings, popularity smoothing, and popularity bootstrapping—to enhance generalizability. Theoretical analysis shows that the bootstrapping loss eliminates the influence of popularity on model learning, making feature embeddings robust to temporal changes. Experiments on real-world datasets confirm TPAB’s superior stability and generalizability under temporal popularity shifts \cite{RS18}.

Tool recommendation systems on the Galaxy platform help researchers expand their scientific workflows with the right tools. In this study, a tool recommendation system based on the Transformer architecture was developed. Its performance was compared with recurrent neural networks (RNN), convolutional neural networks (CNN), and fully connected neural networks (DNN). The results show that the Transformer has faster convergence time, lower model usage time, higher generalization ability to unseen data, and improved prediction accuracy for infrequently used tools. The trained model can be easily built and deployed on any Galaxy server by combining the weights of the Transformer layers and the tool mapping information. After processing the predictions, invalid tools are removed, and the list of tools is sorted by usage in the past year to display the most frequently used tools at the top. This recommender system enables researchers to discover high-quality tools more efficiently and enhance exploratory data analysis in Galaxy \cite{RS19}.

\section{MedicalRec}
\begin{figure*}[!hptb]
\centering
\begin{center}
\includegraphics[width=1\textwidth ]{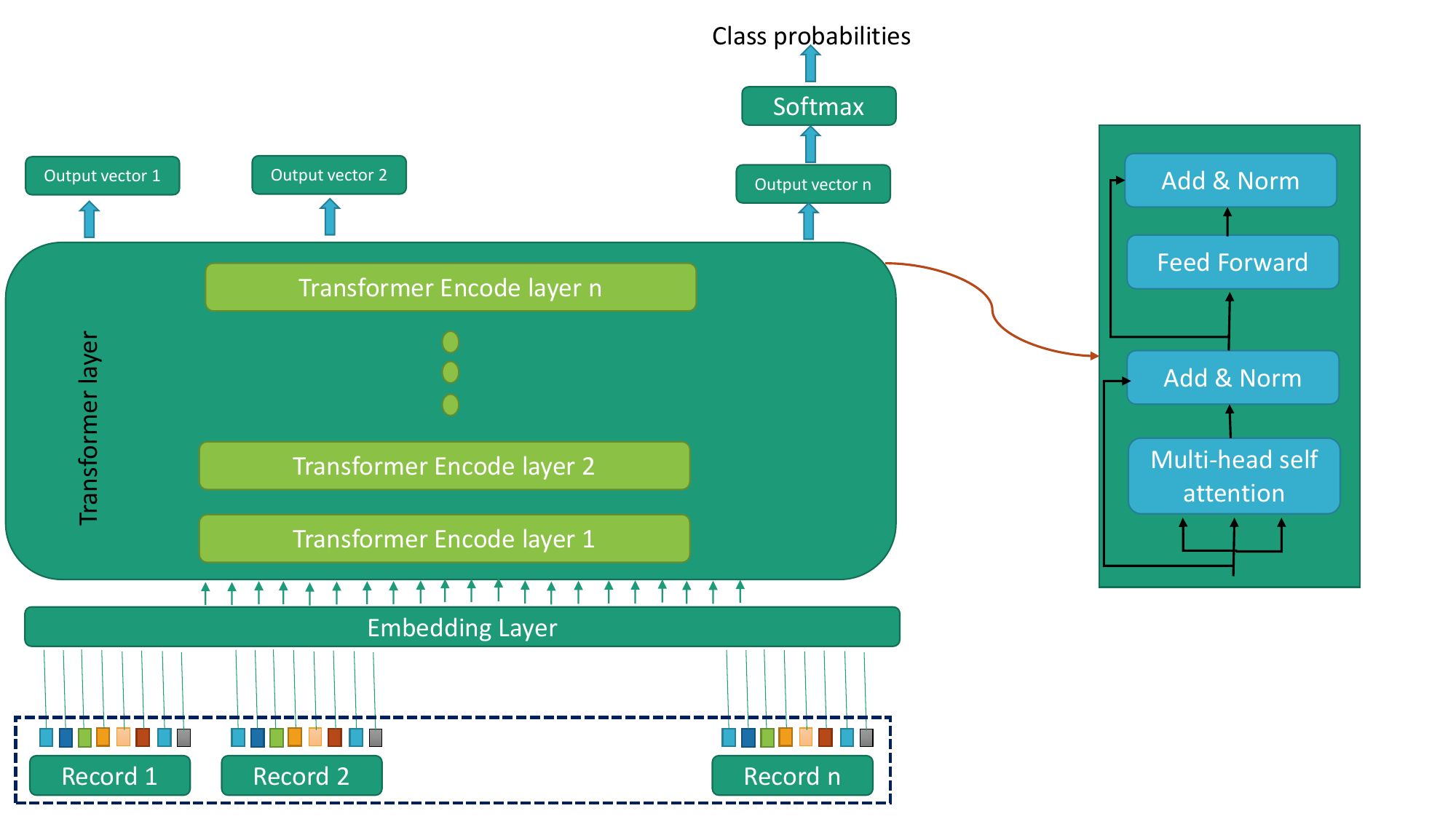}
\caption{MedicalRec architecture.}
\label{Medical}
\end{center}
\end{figure*}
In this section, the goal is to apply a Transformer recommender to the data. The general scheme of MedicalRec approach is shown in Figure \ref{Medical}. In this architecture, the embedding layer learns a representation of the inputs and converts this representation into continuous vectors, or “embeddings.” These embeddings capture the meaning of the input features and provide comprehensive information for subsequent layers, enabling them to perform more accurately. The resulting embeddings are then passed to the Transformer Layer for further processing. The transform layer consists of a set of 12 transform blocks, each with 12 self-attention heads. In this layer, the self-attention heads allow the model to weigh the importance of each item in an input sequence relative to the others. Finally, in the last layer, a SoftMax layer is used to rank all potential recommendations probabilistically.

However, before applying it, some preprocessing is required to prepare the raw inputs for the Transformer model. The dataset contains the following information: \#Sample, \#Train, \#Test, \#Validation, Image Width, Image Height, \#Class, Accuracy, Precision, Recall, F1 Score, AUC, and Model. Each row represents a model (such as CNN, Transformer, or Ensemble) along with its performance metrics (e.g., Accuracy, Precision). This data can be used to recommend appropriate models based on performance metrics, number of classes, or image size. Some columns (such as AUC, Sensitivity, Specificity) have a large amount of missing data (0.0), which is challenging. The dataset is designed for medical image classification; however, for a recommender system, the data needs to be converted into a format such as user-item or criterion-model interactions. Hence, we make a few assumptions:

\begin{enumerate}
   \specialitem{editorOrange}{U} \textcolor{editorOrange}{User:}   We assume that each row of the dataset is not equivalent to a "user", but rather a specific scenario or task (e.g., 7-class or 2-class classification).
  \specialitem{editorOrange}{I} \textcolor{editorOrange}{Item:}   Models (e.g., ResNet-50, VGG-16, Swin-V2) are the items to be recommended.
  \specialitem{editorOrange}{S} \textcolor{editorOrange}{Score:}   We use accuracy as the scoring metric. 
  \specialitem{editorOrange}{G} \textcolor{editorOrange}{ Goal:} For a specific task (e.g., number of classes or image size), recommend the best models according to the performance metrics. 
\end{enumerate}

Since the task is a sequential recommendation, the numeric columns (e.g., Accuracy, Precision, Recall, F1) can be used directly. However, the text columns (Model) are converted into embeddings, which utilize BERT for this purpose. For simplicity, we assume that the primary recommendation criterion is accuracy, but we can also consider a combination of criteria (e.g., a weighted average of Accuracy, F1, Precision, and Recall). We also fill in the missing data (0.0) with the column average. The process of applying the Transformer to these sequences is explained below.

In a sequential recommendation task, let $U={u1,u2,...,u_(|U|)}$ represent a set of features, $V={v1,v2,...,v_{|V|}}$ be a set of models. For each features $u \in U$, the interaction history is represented as an ordered sequence $S_u=[v_1^((u)),...,v_t^((u)),...,v_(n_u)^((u))]$ , where $v_t^((u))\in V$ is the item interacted at time $t$ and $n_u$ is the length of the interaction sequence. The task is to predict the interacted item at time $n_u+1$ given $S_u$ formulated as:
\begin{equation}
    p(v_{n_u+1}^u=v|S_u)
\end{equation}
BERT4Rec addresses this problem utilizing the Bidirectional Encoder Representations from Transformers (BERT) framework for the sequential recommendation task \cite{M1}. Contrary to RNNs and CNNs, the self-attention mechanism enables the BERT4Rec model to capture dependencies across arbitrary distances, resulting in a global receptive field and parallel computation. Additionally, unlike prior unidirectional architectures, its bidirectional encoder model provides richer representations.

Positional embeddings $P \in R^(N×d)$ are added to the input item embeddings to preserve the temporal information. The input representation $h_i^0$ for a given item $v_i$ is formulated as:
\begin{equation}
    h_i^0=e_{v_i}+p_i
\end{equation}
where $e_{v_i}\in E$ is the d-dimensional embedding for item $v_i$, and $p_i \in P$ is the d-dimensional learnable positional embedding for position $i$. If a sequence length exceeds a predefined maximum length of $N$, only the most recent $N$ items are retained, therefore trimming the input sequence. The combined embeddings serve as the input to the transformer stack.
\begin{equation}
    [v_{t-N+1}^u,...,v_t] \text{if t}>N
\end{equation}

The encoder comprises $L$ bidirectional transformer layers, each consisting of two sublayers: A multi-head self-attention (MHSA) and a position-wise feed-forward network (PFFN). Given an input sequence of length $t$, for each position $i$ at each layer $l$, the model iteratively computes the hidden representation $h_i^l\in R^d$, and stacks these together into matrix $H^l\in R^{t×d}$ simultaneously. Stacking $L$ layers yields contextualized hidden states for all positions \cite{M2}.

Self-attention allows every item representation to attend directly to all other items in the sequence. Specifically, the input matrix $H^l$ is projected linearly into queries (Q), keys (K), and values (V) in h subspaces. Attention weights are computed as (as the scaled dot product of the queries and keys, normalized with softmax, and then used to weight the values):
\begin{equation}
    Attention(Q,K,V)=softmax(\frac{QK^T}{\sqrt{d/h}})v
\end{equation}
The temperature $\sqrt{d/h}$ softens the distribution and stabilizes gradients \cite{M3}\cite{M4}. Multiple attention heads are computed in parallel with independent projections and concatenated before a final linear projection:
\begin{equation}
    MHSA(H^l )=Concat[head_1;head_2;…;head_h]W^O
\end{equation}
\begin{equation}
    head_i=Attention(H^l W_i^Q,H^l W_i^K,H^l W_i^V)
\end{equation}
Where $W_i^Q\in R^{d×d/h}, W_i^K\in R^{d×d/h}, W_i^V\in R^{d×d/h}$, and $W_i^O\in R^{d×d}$ are learnable parameters. This enables the model to capture diverse dependency patterns across positions.

The second sublayer is a feed-forward network that transforms each position independently through a two-layer perceptron with the Gaussian error linear unit (GELU) activation function:
\begin{equation}
    PFFN(H^l )=[FFN(h_1^l )^T;…;FFN(h_t^l )^T ]^T
\end{equation}
\begin{equation}
FFN(x)=GELU(xW^1+b^1) W^2+b^2  
\end{equation}

Where $W^1\in R^{d×4d}, W^2\in R^{4d×d}$, and $b^1\in R^{4d}, b^2\in R^d$ are biases. The GELU function is defined as:
\begin{equation}
   GELU(x)=x\Phi(x) 
\end{equation}
Where $\Phi(x)$ is the cumulative distribution function of the standard Gaussian distribution. This activation provides a smoother gradient flow than the rectified linear unit (ReLU), improving optimization \cite{M4}.

Each sublayer output is wrapped with residual connections \cite{M5}, layer normalization \cite{M6}, and dropout \cite{M7}:
\begin{equation}
  LN(x+Dropout(sublayer(x)))  
\end{equation}

Where sublayer(x) is the function of each sublayer itself and LN is the layer normalization function.
The overall BERT4Rec hidden representation refinement of each layer is:
\begin{equation}
    H^l=Trm(H^(l-1) ),\forall_i  \in [1,…,L]
\end{equation}
\begin{equation}
Trm(H^(l-1) )=LN(A^(l-1)+Dropout(PFFN(A^(l-1))))
\end{equation}
\begin{equation}
A^{(l-1)}=LN(H^{(l-1)}+Dropout(MH(H^{(l-1)})))
\end{equation}
producing the final output for all items of all layers represented as $H^L$.
During training, a fixed proportion of input items is randomly replaced by a mask token, and the model’s prediction at those positions is passed through a two-layer feed-forward network with GELU activation:
\begin{equation}
    P(v)=softmax(GELU(h_t^L W^P+b^P ) E^T+b^O)
\end{equation}

where $W^P$ is the learnable projection matrix, tied to the input embedding matrix $E\in R^(|V|×d)$ to reduce parameter count and regularize training. ($b^P$ and $b^O$ are bias terms). The loss is the average negative log-likelihood over masked positions.

\section{MedicalRec-Bench}

In this study, a comprehensive dataset was compiled from more than 3,000 scientific articles spanning diverse topics in machine learning and recommender systems. The aim was to systematically extract relevant metadata and performance metrics from each article to enable thorough evaluation and comparative analysis.
The following fields were consistently retrieved from each article:
\begin{itemize}
    \item  \textbf{Reference:} Bibliographic citation for the article.
\item  \textbf{Title:} The article’s formal title indicating the research focus.
\item  \textbf{Abstract:} A brief summary of the research objectives, methodology, and results.
\item  \textbf{Keywords:} Core terms defining the articles domain.
\item  \textbf{Dataset:} Description or name of the dataset used.
\item  \textbf{Sample Size:} Number of samples employed in experiments.
\item  \textbf{Train/Test/Validation Split:} Data partitions used for model training, evaluation, and validation.
\item  \textbf{K-fold Cross-validation:} Usage of k-fold validation techniques when applicable.
\item  \textbf{Image Dimensions (Width, Height, Channel):} Image-specific attributes where relevant.
\item \textbf{ Number of Classes:} Classes involved in classification tasks.
\item  \textbf{Domain:} Application or research area domain.
\item  \textbf{Performance Metrics:} Including accuracy, precision, recall, F1-score, area under the ROC curve (AUC), sensitivity, and specificity.
\item  \textbf{Model Architecture:} The type or name of the machine learning model used Defined according to standard metrics described in the source article, such as Precision@K, Recall@K, Mean Reciprocal Rank (MRR), and Mean Average Precision (MAP), these fields provide a robust framework for assessing the quality and relevance of recommendations. For example, Precision@K evaluates the proportion of relevant items within the top-K
recommendations, and MRR measures the rank of the first relevant item.
\end{itemize}
Due to some articles testing multiple models, stable fields such as title, abstract, and keywords were fixed, whilst performance metrics adapted per model. Additionally, for studies utilizing multiple datasets, dataset-dependent fields like sample size and data splits varied accordingly.
This structured data extraction approach, exemplified by the MedicalRec framework, offers a systematic methodology suitable for meta-analyses and trend evaluations across a wide array of research contributions.

\begin{figure}[!hptb]
\centering
\begin{center}
\includegraphics[width=0.5\textwidth ]{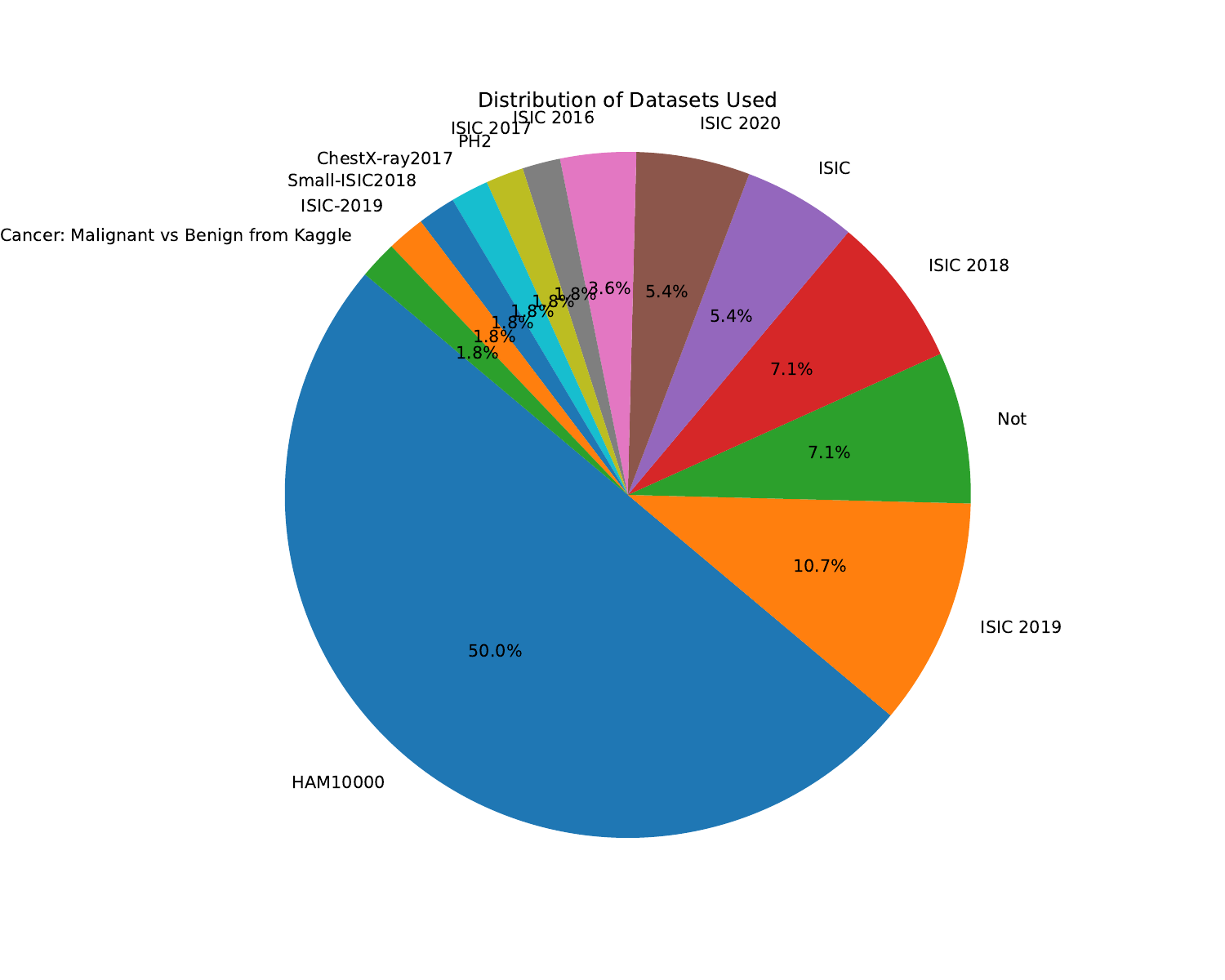}
\caption{Distribution of Datasets Used.}
\label{fig:Medical1}
\end{center}
\end{figure}

\begin{figure}[!hptb]
\centering
\begin{center}
\includegraphics[width=0.5\textwidth ]{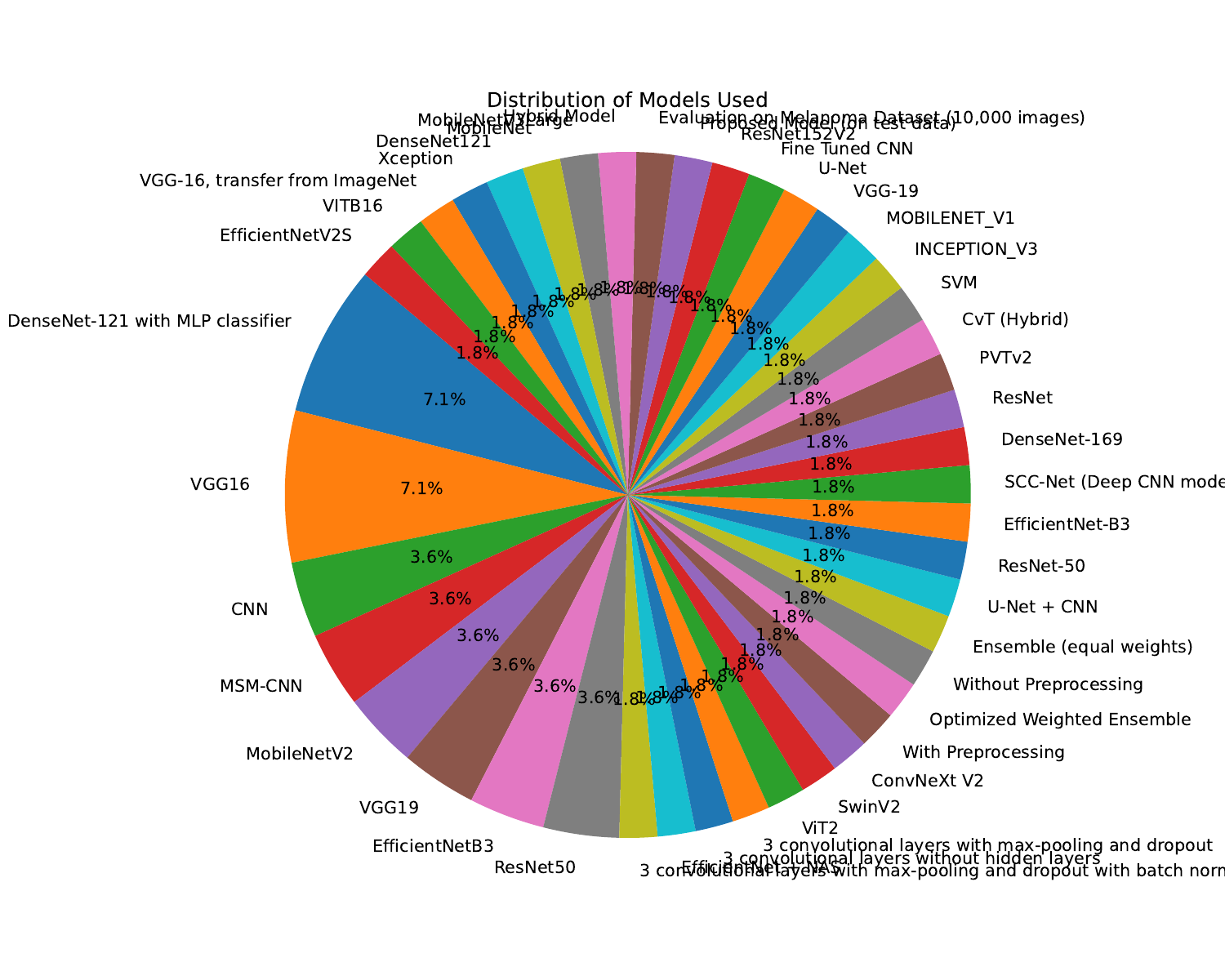}
\caption{Distribution of Models Used.}
\label{fig:Medical2}
\end{center}
\end{figure}

\begin{figure}[!hptb]
\centering
\begin{center}
\includegraphics[width=0.5\textwidth ]{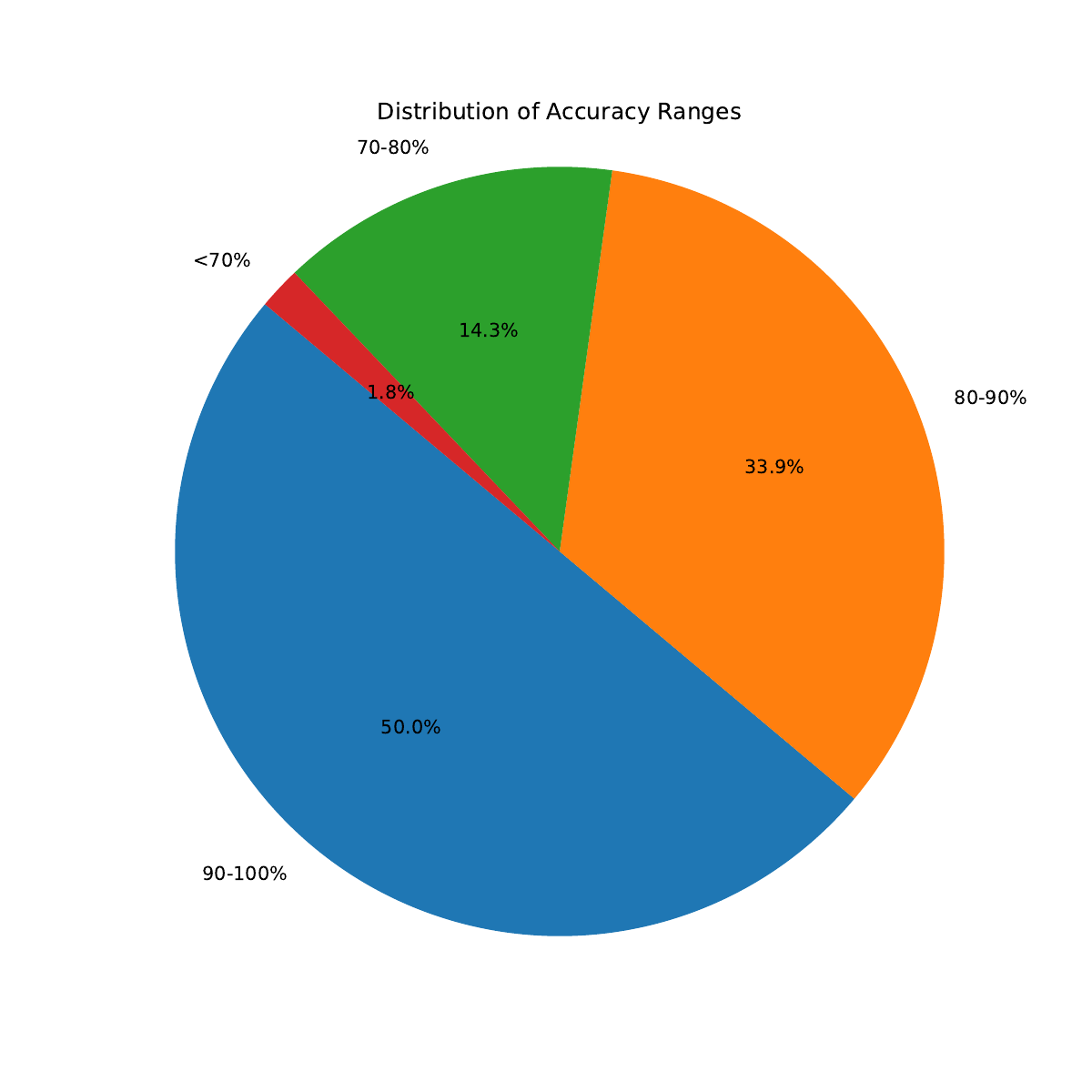}
\caption{Distribution of Accuracy Ranges.}
\label{fig:Medical3}
\end{center}
\end{figure}

Some analytical graphs of the collected data are shown in Figures \ref{fig:Medical1}, \ref{fig:Medical2}, an \ref{fig:Medical3}.


\section{Result}
In this section, we evaluate the proposed model on the MedicalRec dataset. For this purpose, 12 baseline models were evaluated: Hybrid\cite{R1}, SVD\cite{R2}, Collaborative Filtering\cite{R3}, Content-Based\cite{R4}, Large Language Model (LLM)\cite{R5}, Sequential\cite{R6}, Green\cite{R7}, Multi-Modal\cite{R8}, Fair\cite{R9}, LSTM\cite{R10}, GRU\cite{R11}, and IndRNN\cite{R12}. These models were tested on four different input combinations, described as follows:

\begin{enumerate}
    \specialitem{editorGreen}{I} \textcolor{editorGreen}{\textbf{MedicalRec I:}}   This dataset includes the following features: Dataset, Sample Size, Train Size, Test Size, and Validation Size.
   \specialitem{editorGreen}{II} \textcolor{editorGreen}{\textbf{MedicalRec II:}} This dataset includes the following features: Dataset, Sample Size, Train Size, Test Size, Validation Size, K-fold Cross-Validation, Width, Height, and Channel.
    \specialitem{editorGreen}{III} \textcolor{editorGreen}{\textbf{MedicalRec III:}} This dataset includes the following features: Dataset, Sample Size, Train Size, Test Size, Validation Size, K-fold Cross-Validation, Width, Height, Channel, Number of Classes, and Domain.
    \specialitem{editorGreen}{IV} \textcolor{editorGreen}{\textbf{MedicalRec IV:}} This dataset includes the following features: Dataset, Sample Size, Train Size, Test Size, Validation Size, K-fold Cross-Validation, Width, Height, Channel, Number of Classes, Domain, Accuracy, Precision, Recall, F1-Score, Area Under the ROC Curve (AUC), Sensitivity, and Specificity.
\end{enumerate}

In recommender systems, evaluation metrics are used to measure the quality of recommendations. To evaluate the proposed model and the baseline models, four metrics — Precision@K, Recall@K, F1@K,  Discounted Cumulative Gain (DCG), and HitRate@K — were employed. Precision@K represents the ratio of relevant items among the top K recommended items. This metric focuses on the accuracy of recommendations and examines how many of the recommended items are truly suitable for the user, defined as $Precision@K=\frac{\text{Number of relevant items in the top K}}{\text{recommendations K}}$. Recall@K represents the ratio of relevant items among the top K items to the total relevant items in the dataset. This metric focuses on the system's ability to find all relevant items and is defined as $Recall@K=\frac{\text{Number of relevant items in the top K recommendations}}{\text{Total number of relevant items in the dataset}}$. Additionally, F1@K is the harmonic mean of Precision@K and Recall@K, providing a balance between accuracy and coverage, and is defined as $F1@K=2×\frac{Precision@K×Recall@K}{Precision@K+Recall@K} $. DCG is different from the introduced metrics. This is a ranking-based metric that evaluates the quality of suggestions based on their position in the suggestion list. Relevant items in higher rankings are given more weight because users tend to pay more attention to items at the beginning of the list, and is defined as $\text{DCG@K} = \sum_{i=1}^{K} \frac{\text{rel}_i}{\log_2(i+1)}$. HitRate@K it focuses on whether the system provides at least one relevant suggestion regardless of the total number of relevant items and is defined as $\text{HitRate@k} = \frac{|\{u \in U : \exists i \in \text{Top-k}(u), \text{rel}_i = 1\}|}{|U|}$.

\subsection{Result of ablation study}
The ablation study with different changes of elements in the proposed model is presented in Table \ref{Table:1R}. The value of F1@100 was considered for evaluation. The results of erosion studies conducted with the experimental dataset were considered for two Case studies:

\begin{enumerate}
   \specialitem{editorBlue}{A} \textcolor{editorBlue}{Case study 1:}  Activation Function: Different activation functions can affect the performance of a model; hence, choosing an optimal activation function is relevant in building an optimal model. Six activation functions, including Linear, PReLU, LeakyReLU, Tanh, ELU, and ReLU, have been tested. It is found that the ReLU activation function yielded the highest result across all the data.
  \specialitem{editorBlue}{D} \textcolor{editorBlue}{Case study 2:}  Dropout. To demonstrate the impact of Dropout on the model, drop rates of 0.5, 0.6, 0.7, 0.8, and 0.9 were used. Considering these values, the proposed model achieved the best results in MedicalRec I with a Drop rate of 0.7, in MedicalRec II with a Drop rate of 0.5, in MedicalRec III with a Drop rate of 0.5, and in MedicalRec IV with a Drop rate of 0.5.
\end{enumerate}

\begin{table*}[]
\centering
\caption{Ablation study regarding layer  activation functions and dropout.}
\begin{tabular}{|lccl|cl|}
\hline
\multicolumn{4}{|c|}{MedicalRec I}                                                                                                                                                         & \multicolumn{2}{c|}{MedicalRec II}                \\ \hline
\multicolumn{1}{|l|}{Case study 1: changing activation function}                            & \multicolumn{1}{c|}{Activation Function} & \multicolumn{1}{c|}{F1@100} & Finding             & \multicolumn{1}{c|}{F1@100} & Finding             \\ \hline
\multicolumn{1}{|l|}{1}                                                                     & \multicolumn{1}{c|}{Linear}              & \multicolumn{1}{c|}{2.78}   & \textcolor[rgb]{0.00,1.00,0.00}{Improved F1@100 }     & \multicolumn{1}{c|}{9.32}   &\textcolor[rgb]{0.00,1.00,0.00}{Improved F1@100 }       \\ \hline
\multicolumn{1}{|l|}{2}                                                                     & \multicolumn{1}{c|}{PRelue}              & \multicolumn{1}{c|}{2.73}   & \textcolor[rgb]{1.00,0.28,0.28}{Not Improved F1@100} & \multicolumn{1}{c|}{9.11}   & \textcolor[rgb]{1.00,0.28,0.28}{Not Improved F1@100} \\ \hline
\multicolumn{1}{|l|}{3}                                                                     & \multicolumn{1}{c|}{LeakyReLu}           & \multicolumn{1}{c|}{2.83}   & \textcolor[rgb]{0.00,1.00,0.00}{Improved F1@100 }       & \multicolumn{1}{c|}{9.34}   & \textcolor[rgb]{0.00,1.00,0.00}{Improved F1@100 }      \\ \hline
\multicolumn{1}{|l|}{4}                                                                     & \multicolumn{1}{c|}{Tanh}                & \multicolumn{1}{c|}{2.84}   & \textcolor[rgb]{0.00,1.00,0.00}{Improved F1@100 }       & \multicolumn{1}{c|}{9.51}   &\textcolor[rgb]{0.00,1.00,0.00}{Improved F1@100 }      \\ \hline
\multicolumn{1}{|l|}{5}                                                                     & \multicolumn{1}{c|}{ELU}                 & \multicolumn{1}{c|}{2.92}   & \textcolor[rgb]{0.00,1.00,0.00}{Improved F1@100 }      & \multicolumn{1}{c|}{9.13}   & \textcolor[rgb]{1.00,0.28,0.28}{Not Improved F1@100} \\ \hline
\multicolumn{1}{|l|}{6}                                                                     & \multicolumn{1}{c|}{ReLu}                & \multicolumn{1}{c|}{2.93}   &                     & \multicolumn{1}{c|}{9.52}   & \textcolor[rgb]{0.00,1.00,0.00}{Improved F1@100 }      \\ \hline
\multicolumn{1}{|l|}{Case study 2: Dropout} & \multicolumn{1}{c|}{Drop rate}           & \multicolumn{1}{c|}{F1@100} & Finding             & \multicolumn{1}{c|}{F1@100} & Finding             \\ \hline
\multicolumn{1}{|l|}{1}                                                                     & \multicolumn{1}{c|}{0.5}                 & \multicolumn{1}{c|}{2.93}   & \textcolor[rgb]{1.00,0.28,0.28}{Not Improved F1@100} & \multicolumn{1}{c|}{9.60}   & \textcolor[rgb]{0.00,1.00,0.00}{Improved F1@100 }       \\ \hline
\multicolumn{1}{|l|}{2}                                                                     & \multicolumn{1}{c|}{0.6}                 & \multicolumn{1}{c|}{2.94}   & \textcolor[rgb]{0.00,1.00,0.00}{Improved F1@100 }      & \multicolumn{1}{c|}{9.53}   & \textcolor[rgb]{1.00,0.28,0.28}{Not Improved F1@100} \\ \hline
\multicolumn{1}{|l|}{3}                                                                     & \multicolumn{1}{c|}{0.7}                 & \multicolumn{1}{c|}{3.00}   & \textcolor[rgb]{0.00,1.00,0.00}{Improved F1@100 }      & \multicolumn{1}{c|}{9.53}   & \textcolor[rgb]{1.00,0.28,0.28}{Not Improved F1@100} \\ \hline
\multicolumn{1}{|l|}{4}                                                                     & \multicolumn{1}{c|}{0.8}                 & \multicolumn{1}{c|}{2.93}   & \textcolor[rgb]{1.00,0.28,0.28}{Not Improved F1@100} & \multicolumn{1}{c|}{9.56}   & \textcolor[rgb]{1.00,0.28,0.28}{Not Improved F1@100} \\ \hline
\multicolumn{1}{|l|}{5}                                                                     & \multicolumn{1}{c|}{0.9}                 & \multicolumn{1}{c|}{2.91}   & \textcolor[rgb]{1.00,0.28,0.28}{Not Improved F1@100} & \multicolumn{1}{c|}{9.59}   & \textcolor[rgb]{1.00,0.28,0.28}{Not Improved F1@100} \\ \hline
\multicolumn{4}{|c|}{MedicalRec III}                                                                                                                                                       & \multicolumn{2}{c|}{MedicalRec IV}                \\ \hline
\multicolumn{1}{|l|}{Case study 1: changing activation function}                            & \multicolumn{1}{c|}{Activation Function} & \multicolumn{1}{c|}{F1@100} & Finding             & \multicolumn{1}{c|}{F1@100} & Finding             \\ \hline
\multicolumn{1}{|l|}{1}                                                                     & \multicolumn{1}{c|}{Linear}              & \multicolumn{1}{c|}{2.74}   & \textcolor[rgb]{0.00,1.00,0.00}{Improved F1@100 }    & \multicolumn{1}{c|}{9.18}   & \textcolor[rgb]{0.00,1.00,0.00}{Improved F1@100 }       \\ \hline
\multicolumn{1}{|l|}{2}                                                                     & \multicolumn{1}{c|}{PRelue}              & \multicolumn{1}{c|}{2.75}   & \textcolor[rgb]{0.00,1.00,0.00}{Improved F1@100 }      & \multicolumn{1}{c|}{9.32}   & \textcolor[rgb]{0.00,1.00,0.00}{Improved F1@100 }      \\ \hline
\multicolumn{1}{|l|}{3}                                                                     & \multicolumn{1}{c|}{LeakyReLu}           & \multicolumn{1}{c|}{2.86}   & \textcolor[rgb]{0.00,1.00,0.00}{Improved F1@100 }       & \multicolumn{1}{c|}{9.44}   & \textcolor[rgb]{0.00,1.00,0.00}{Improved F1@100 }       \\ \hline
\multicolumn{1}{|l|}{4}                                                                     & \multicolumn{1}{c|}{Tanh}                & \multicolumn{1}{c|}{2.83}   & \textcolor[rgb]{1.00,0.28,0.28}{Not Improved F1@100} & \multicolumn{1}{c|}{9.52}   & \textcolor[rgb]{0.00,1.00,0.00}{Improved F1@100 }       \\ \hline
\multicolumn{1}{|l|}{5}                                                                     & \multicolumn{1}{c|}{ELU}                 & \multicolumn{1}{c|}{2.94}   & \textcolor[rgb]{0.00,1.00,0.00}{Improved F1@100 }       & \multicolumn{1}{c|}{9.71}   & \textcolor[rgb]{0.00,1.00,0.00}{Improved F1@100 }       \\ \hline
\multicolumn{1}{|l|}{6}                                                                     & \multicolumn{1}{c|}{Relu}                & \multicolumn{1}{c|}{2.97}   & \textcolor[rgb]{0.00,1.00,0.00}{Improved F1@100 }       & \multicolumn{1}{c|}{9.73}   & \textcolor[rgb]{0.00,1.00,0.00}{Improved F1@100 }       \\ \hline
\multicolumn{1}{|l|}{Case study 2: Dropout} & \multicolumn{1}{c|}{Drop rate}           & \multicolumn{1}{c|}{F1@100} & Finding             & \multicolumn{1}{c|}{F1@100} & Finding             \\ \hline
\multicolumn{1}{|l|}{1}                                                                     & \multicolumn{1}{c|}{0.5}                 & \multicolumn{1}{c|}{3.00}   & \textcolor[rgb]{0.00,1.00,0.00}{Improved F1@100 }       & \multicolumn{1}{c|}{9.80}   & \textcolor[rgb]{0.00,1.00,0.00}{Improved F1@100 }      \\ \hline
\multicolumn{1}{|l|}{2}                                                                     & \multicolumn{1}{c|}{0.6}                 & \multicolumn{1}{c|}{2.65}   & \textcolor[rgb]{1.00,0.28,0.28}{Not Improved F1@100} & \multicolumn{1}{c|}{9.46}   & \textcolor[rgb]{0.00,1.00,0.00}{Improved F1@100 }     \\ \hline
\multicolumn{1}{|l|}{3}                                                                     & \multicolumn{1}{c|}{0.7}                 & \multicolumn{1}{c|}{2.86}   & \textcolor[rgb]{1.00,0.28,0.28}{Not Improved F1@100} & \multicolumn{1}{c|}{9.45}   & \textcolor[rgb]{1.00,0.28,0.28}{Not Improved F1@100} \\ \hline
\multicolumn{1}{|l|}{4}                                                                     & \multicolumn{1}{c|}{0.8}                 & \multicolumn{1}{c|}{2.88}   & \textcolor[rgb]{1.00,0.28,0.28}{Not Improved F1@100} & \multicolumn{1}{c|}{9.76}   & \textcolor[rgb]{1.00,0.28,0.28}{Not Improved F1@100} \\ \hline
\multicolumn{1}{|l|}{5}                                                                     & \multicolumn{1}{c|}{0.9}                 & \multicolumn{1}{c|}{2.98}   & \textcolor[rgb]{1.00,0.28,0.28}{Not Improved F1@100} & \multicolumn{1}{c|}{9.77}   & \textcolor[rgb]{1.00,0.28,0.28}{Not Improved F1@100} \\ \hline
\end{tabular}
\label{Table:1R}
\end{table*}

Table \ref{Table:2R} shows the results obtained by the baseline and proposed approaches on different data combinations. To scientifically examine the presented table in relation to recommender systems, we analyze the data from the perspective of evaluation criteria, algorithm performance, and scientific concepts related to recommender systems. The table includes the Discounted Cumulative Gain (DCG) measure at different points (D@5, D@10, D@20, D@50, D@75) for several recommender models. The values of D@5, D@10, and so on, indicate the model's performance in recommending the top k items. The higher the DCG value, the more successful the model is in providing more relevant recommendations. The DCG values increase with increasing k (from 5 to 75), which is expected, as the number of recommended items increases, the probability of including relevant items also increases, as compared to model performance. The table includes 13 recommender models, comprising classical approaches such as collaborative filtering and content-based filtering, as well as deep learning-based models like LSTM, GRU, and IndRNN, and hybrid and novel models, including multi-modal, LLM, and recommendation models. The following analysis examines the performance of the models.

\textbf{MedicalRec I:} The proposed model achieved the highest scores in all metrics (0.1018 at D@5 to 0.1970 at D@75). This model significantly outperformed the other models at all evaluation points. The Fair model is the second-best model, with scores ranging from 0.1080 (D@5) to 0.1916 (D@75). This model outperforms the proposed model at D@5 (0.1080 vs. 0.1018), but lags behind the proposed model at later points (D@10 onwards). This may indicate that the model prioritizes improving accuracy in the early ranks, which is particularly useful for systems where users typically only pay attention to the first few items. Fair models are typically designed to mitigate bias in recommendations, and this strong performance suggests that the model has struck a good balance between accuracy and fairness. In deep learning-based models (LSTM, GRU, IndRNN), the GRU model performs best among the three models (0.0986 at D@5 to 0.1791 at D@75). LSTM and IndRNN perform similarly, but are slightly weaker than GRU. Recurrent network (RNN)-based models are suitable for modelling sequential patterns in data (such as user interaction history). The better performance of GRU may be due to its ability to handle vanishing gradients and learn long-term patterns. These models perform worse than the proposed model and Fair, which may be due to lower complexity or a lack of use of multimodal information.

    The classical models (Collaborative Filtering, Content-Based, and SVD) also showed poor performance. The Collaborative Filtering model had the weakest performance (0.0503 in D@5 to 0.1037 in D@75). Content-Based performed better than Collaborative Filtering (0.0887 in D@5 to 0.1765 in D@75), while the SVD model recorded average performance (0.0769 in D@5 to 0.1671 in D@75). Collaborative Filtering performs poorly in cases of data scarcity or issues such as a cold start. Content-Based performs better by utilizing the content features of items, such as text descriptions or labels, but still lags behind hybrid models. As a matrix factorization-based method, SVD has mediocre performance and likely lags behind deep learning models due to its limitations in modelling more complex patterns.

    The hybrid and novel models (Hybrid, LLM, Sequential, Multi-modal, Green) have obtained better results than the other models. The Hybrid and Multi-modal models had similar and close performance to the proposed model (0.0940 in D@5 for both). The three models, LLM, Sequential, and Green, recorded similar performance (around 0.0922–0.0926 in D@5). Hybrid models (Hybrid, Multi-modal) perform better than the classical models by combining information. LLM  performs well due to its ability to understand semantics, but it does not reach the proposed or Fair model. Sequential focuses on sequential data, and its performance is comparable to that of LLM and Green. The proposed model, with excellence in all metrics, represents a robust and possibly hybrid approach that can be suitable for various recommender scenarios. The Fair model’s strong performance in D@5 indicates that it is suitable for applications that require high accuracy in the first few recommendations.

\textbf{Medical Rec II:} The DCG values in this dataset (0.0228 to 0.09094) are significantly lower than those in Medical Rec I (0.0503 to 0.1970). As k increases from D@5 to D@100, the DCG values increase, which makes sense since the probability of including relevant items increases with the number of recommended items. The Proposed Model achieves the highest scores in most of the metrics (0.0406 at D@5, 0.0412 at D@10, 0.0534 at D@20, 0.0786 at D@50, 0.0855 at D@75, 0.09094 at D@100). The model consistently performs best at all evaluation points except D@10, which is slightly lower than the fair model. This indicates a strong ability to rank related items, especially in longer lists. The Fair model achieved the second-best performance with scores of 0.0402 (D@5), 0.0484 (D@10), 0.0590 (D@20), 0.0760 (D@50), 0.0844 (D@75), 0.0908 (D@100). The model performs very well at D@5 and D@10 and even outperforms the proposed model at D@10 (0.0484). This suggests that the model is optimized for short recommendations (the first 5 or 10 items). The LLM, Sequential, Green, Multi-modal models perform similarly and rank in the middle. Close values indicate the use of similar architectures or the same input data. In deep learning-based models (LSTM, GRU, IndRNN), the GRU and IndRNN approaches outperform LSTM, which may be due to their enhanced ability to handle long-term dependencies in sequential data. However, these models are weaker compared to the proposed and Fair models. Also in classical models (Hybrid, SVD, Collaborative Filtering, Content-Based), the Collaborative Filtering and SVD approaches perform the weakest, which may be due to cold start problems or limitations in interactive data. Also, Content-Based and Hybrid perform better. Overall, it can be concluded that the Proposed Model performs best with scores ranging from 0.0406 (D@5) to 0.09094 (D@100). The Fair model is very suitable for short recommendations with a score of 0.0484 at D@10. On the other hand, in classical models, Collaborative Filtering and SVD perform the weakest. Also, in deep learning models, GRU and IndRNN perform better than LSTM, but are weaker than the proposed model.

\textbf{Medical Rec II:} The DCG values in this dataset (0.0228 to 0.09094) are significantly lower than those in Medical Rec I (0.0503 to 0.1970). As k increases from D@5 to D@100, the DCG values increase, which makes sense since the probability of including relevant items increases with the number of recommended items. The Proposed Model achieves the highest scores in most of the metrics (0.0406 at D@5, 0.0412 at D@10, 0.0534 at D@20, 0.0786 at D@50, 0.0855 at D@75, 0.09094 at D@100). The model consistently performs best at all evaluation points except D@10, which is slightly lower than the fair model. This indicates a strong ability to rank related items, especially in longer lists. The Fair model achieved the second-best performance with scores of 0.0402 (D@5), 0.0484 (D@10), 0.0590 (D@20), 0.0760 (D@50), 0.0844 (D@75), 0.0908 (D@100). The model performs very well at D@5 and D@10 and even outperforms the proposed model at D@10 (0.0484). This suggests that the model is optimized for short recommendations (the first 5 or 10 items). The LLM, Sequential, Green, Multi-modal models perform similarly and rank in the middle. Close values indicate the use of similar architectures or the same input data. In deep learning-based models (LSTM, GRU, IndRNN), the GRU and IndRNN approaches outperform LSTM, which may be due to their enhanced ability to handle long-term dependencies in sequential data. However, these models are weaker compared to the proposed and Fair models. Also in classical models (Hybrid, SVD, Collaborative Filtering, Content-Based), the Collaborative Filtering and SVD approaches perform the weakest, which may be due to cold start problems or limitations in interactive data. Also, Content-Based and Hybrid perform better. Overall, it can be concluded that the Proposed Model performs best with scores ranging from 0.0406 (D@5) to 0.09094 (D@100). The Fair model is very suitable for short recommendations with a score of 0.0484 at D@10. On the other hand, in classical models, Collaborative Filtering and SVD perform the weakest. Also, in deep learning models, GRU and IndRNN perform better than LSTM, but are weaker than the proposed model.

\textbf{Medical IV:} The DCG values in this table (0.0397–0.0920) are significantly lower than those in Medical III (0.1012–0.2101). The two models, Proposed Model and IndRNN, perform best. Most models proliferate from D@5 to D@20, but the growth slows down from D@50 onwards. Proposed Model scores the highest in most metrics (0.0512 at D@5, 0.0511 at D@10, 0.0687 at D@20, 0.0867 at D@50, 0.0901 at D@75, 0.0920 at D@100). This model leads at all evaluation points (especially D@20 onwards). The significant superiority at D@20 indicates a strong ability to rank related items in medium and long lists. The IndRNN performs very well at D@5 and D@10 and is only 0.0001 behind the proposed model at D@100. The IndRNN architecture achieves strong performance due to its ability to model long-term dependencies in sequential data. The middle group, which includes Hybrid, SVD, Collaborative Filtering, Content-Based, LLM, Sequential, Green, Multi-modal, Fair, LSTM, and GRU, performs very closely (with a difference of less than 1\% in most metrics), indicating the use of similar architectures or input data. GRU and Fair perform slightly better than the rest of the group at D@5 (0.0408 and 0.0405). Classical models, such as Collaborative Filtering, SVD, and Content-Based, perform similarly but slightly worse than deep learning models, including GRU and IndRNN. Next, LLM and Sequential perform the worst at D@5, which may be due to limitations in processing the initial data or a lack of optimization for short lists. However, at D@100, they are close to the middle group. LLM may perform worse on small or sparse datasets due to its high complexity. Sequential models are also likely limited in modelling short-term patterns.

\begin{figure*}
\centering
\subfloat[ MedicalRec I]{\includegraphics[width=0.5\textwidth]{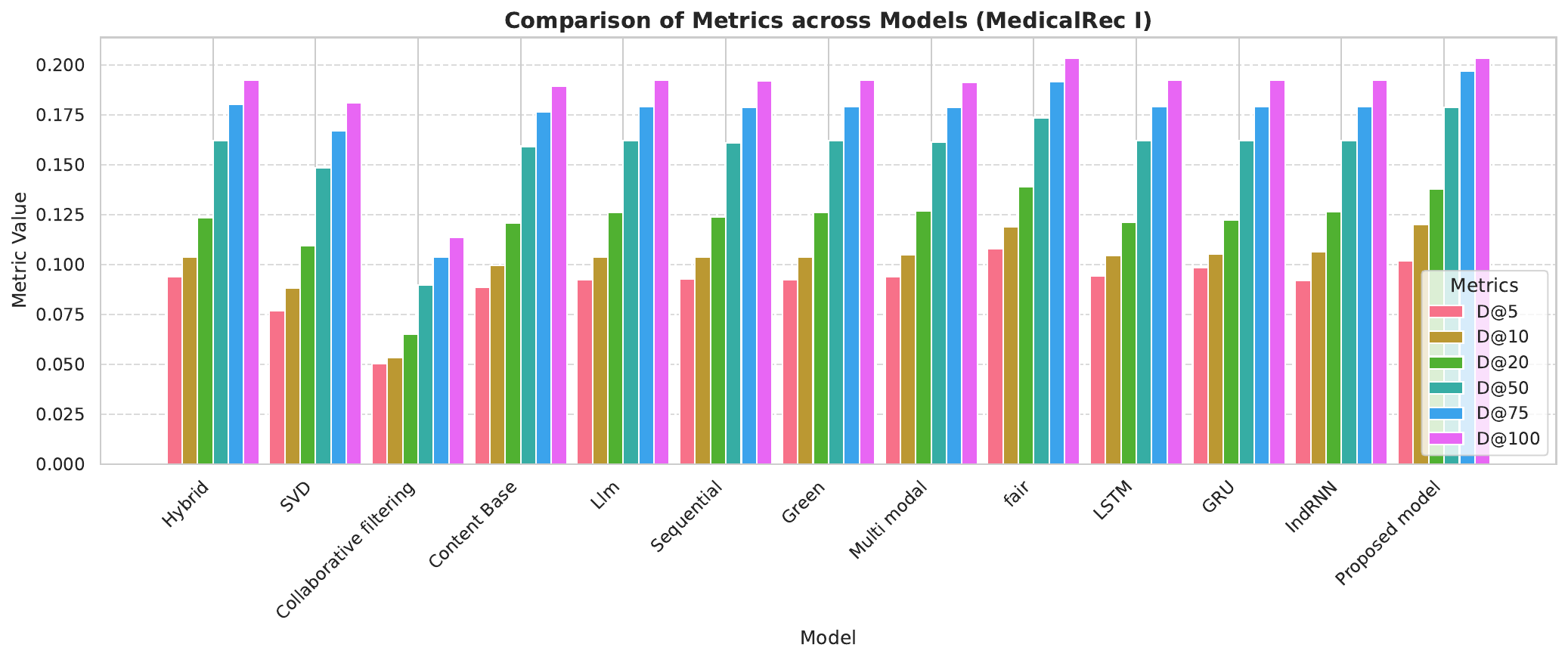}}
\subfloat[MedicalRec II]{\includegraphics[width=0.5\textwidth]{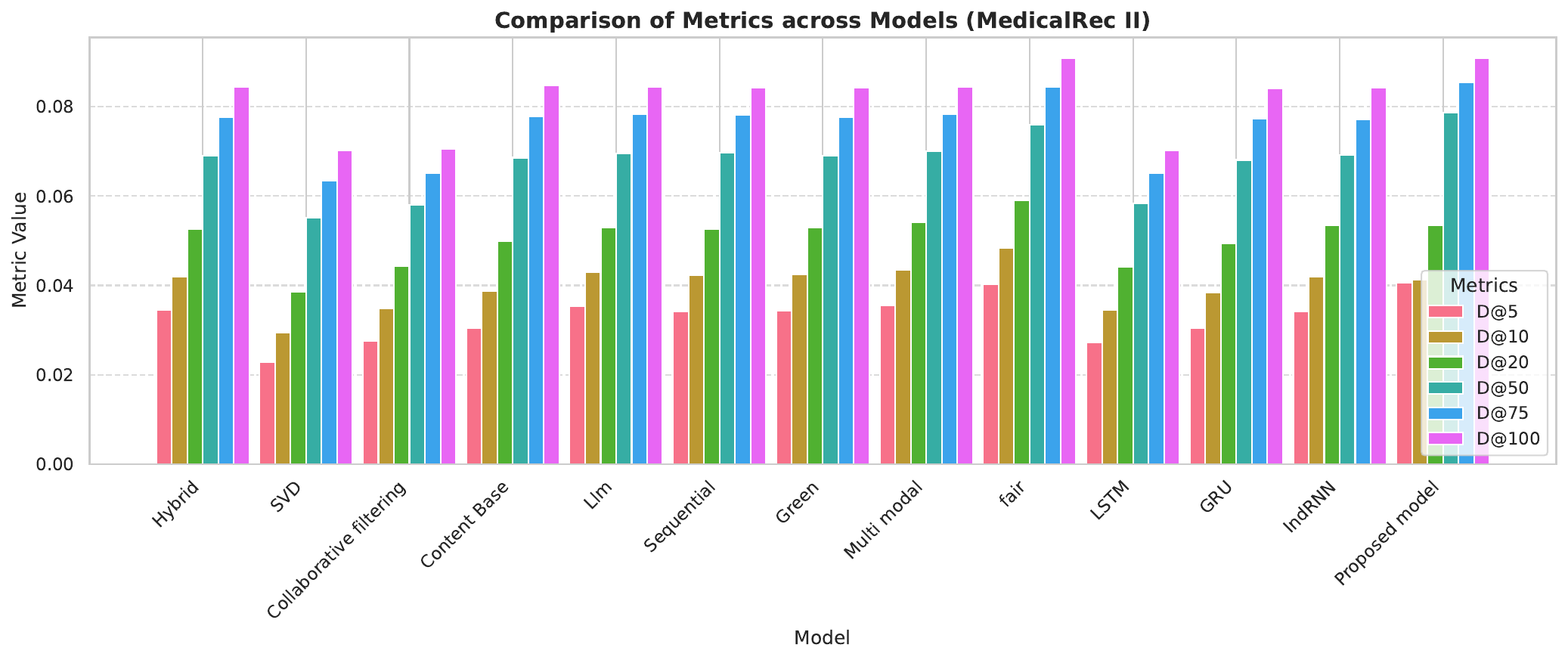}}\\
\subfloat[MedicalRec III]{\includegraphics[width=0.5\textwidth]{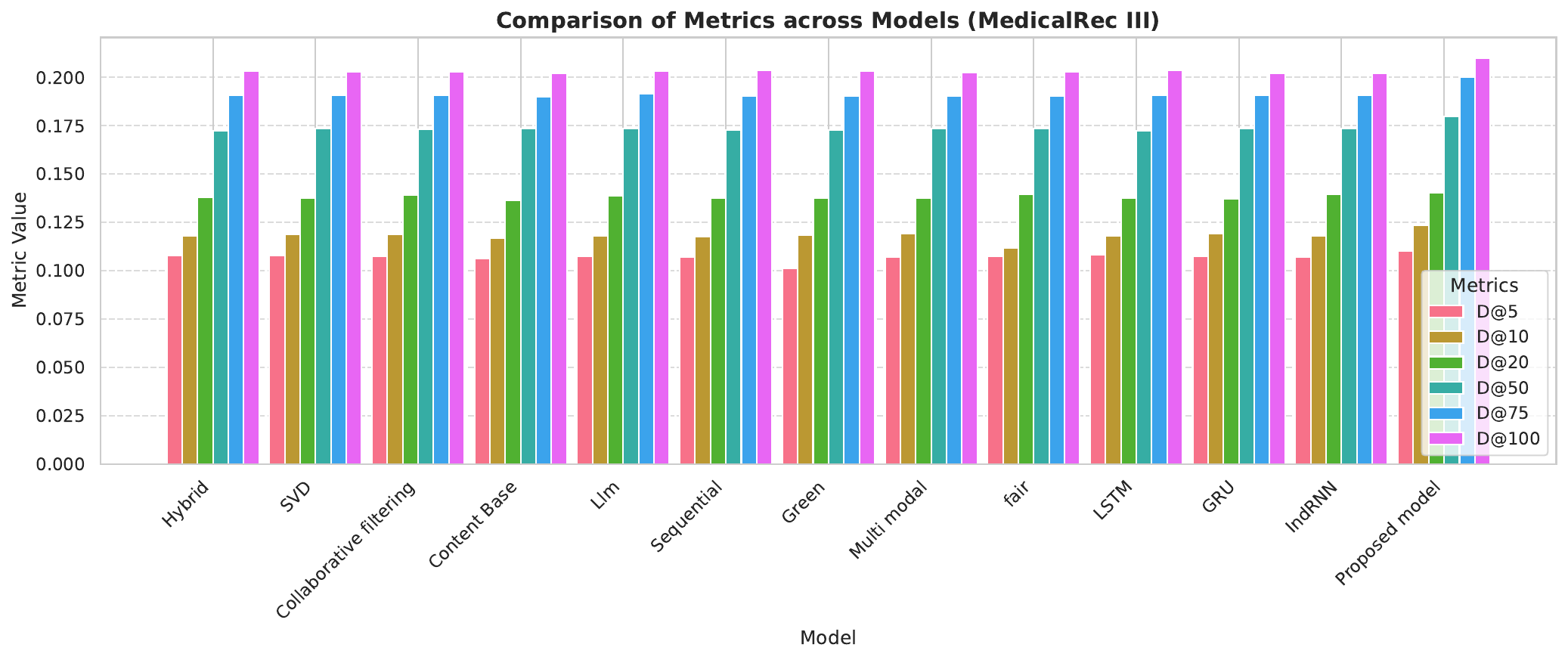}}
\subfloat[MedicalRec IV]{\includegraphics[width=0.5\textwidth]{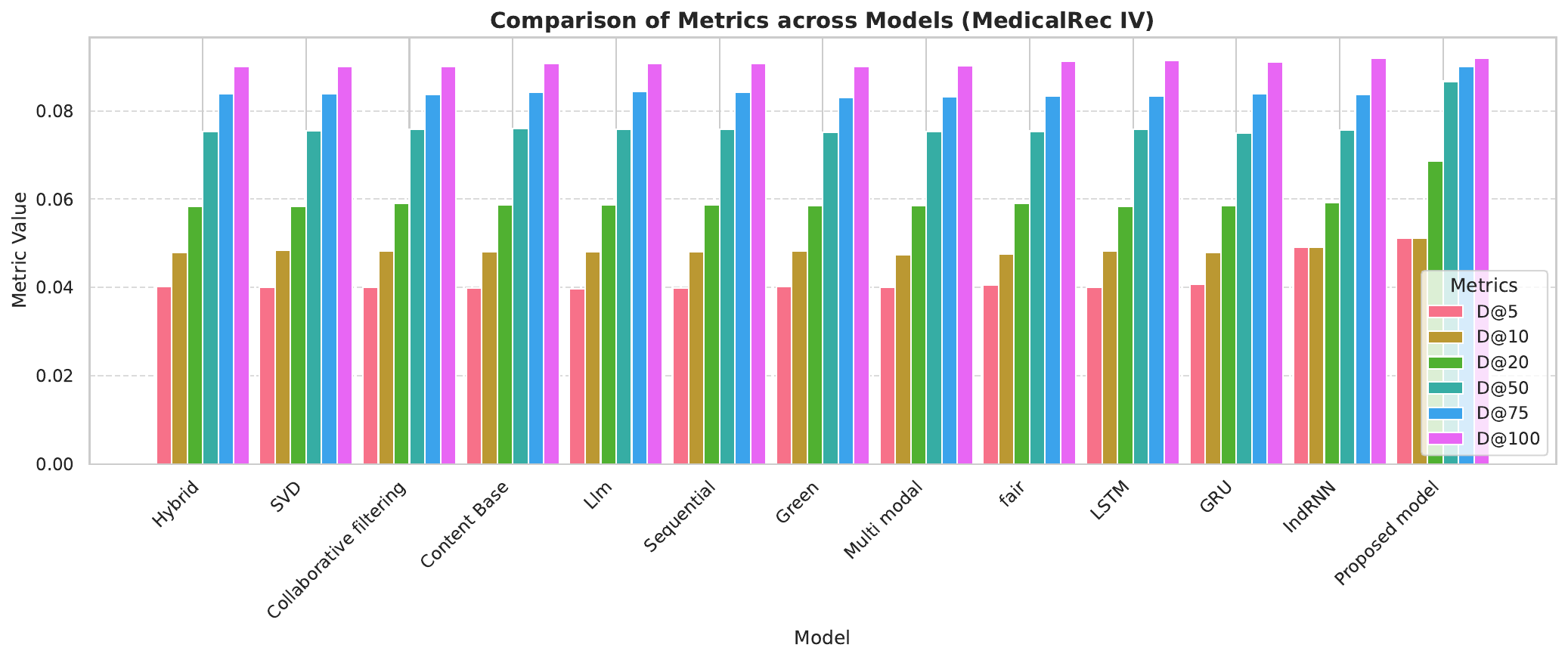}}\\
\caption{Comparison of Metrics across Models.}
\label{fig:fig10}
\end{figure*}

All models exhibit a significant increase in DCG from D@5 to D@20 (e.g., the Proposed Model increases from 0.0512 to 0.0687, a 34\% increase). This suggests that the models can identify more relevant items within the first 20 items. From D@50 to D@100, the DCG growth slows down (e.g., the Proposed Model Increases from 0.0867 to 0.0920, a 6\% increase). This indicates a decrease in efficiency on longer lists, as additional items are less relevant. At D@100, most models have very close scores (0.0900–0.0919), except for the Proposed Model, which has a score of 0.0920. This indicates that on long lists, the difference between models decreases, possibly due to saturation of relevant information or data limitations. The comparative chart resulting from this analysis is shown in Figure \ref{fig:fig10}.

\begin{table*}[]
\caption{Results obtained by the proposed approach and baseline approaches on the MedicalRec I, MedicalRec II, MedicalRec III and MedicalRec IV datasets. Results are reported for the Discounted Cumulative Gain (DCG).}
\begin{tabular}{|l|cccccc|cccccc|}
\hline
Dataset                 & \multicolumn{6}{c|}{MedicalRec I}                                                                                                                            & \multicolumn{6}{c|}{MedicalRec II}                                                                                                                            \\ \hline
Measure\%               & \multicolumn{1}{c|}{D@5}    & \multicolumn{1}{c|}{D@10}   & \multicolumn{1}{c|}{D@20}   & \multicolumn{1}{c|}{D@50}   & \multicolumn{1}{c|}{D@75}   & D@100  & \multicolumn{1}{c|}{D@5}    & \multicolumn{1}{c|}{D@10}   & \multicolumn{1}{c|}{D@20}   & \multicolumn{1}{c|}{D@50}   & \multicolumn{1}{c|}{D@75}   & D@100   \\ \hline
Hybrid                  & \multicolumn{1}{c|}{0.0940} & \multicolumn{1}{c|}{0.1037} & \multicolumn{1}{c|}{0.1234} & \multicolumn{1}{c|}{0.1621} & \multicolumn{1}{c|}{0.1804} & 0.1925 & \multicolumn{1}{c|}{0.0345} & \multicolumn{1}{c|}{0.0419} & \multicolumn{1}{c|}{0.0527} & \multicolumn{1}{c|}{0.0690} & \multicolumn{1}{c|}{0.0777} & 0.0845  \\ \hline
SVD                     & \multicolumn{1}{c|}{0.0769} & \multicolumn{1}{c|}{0.0881} & \multicolumn{1}{c|}{0.1093} & \multicolumn{1}{c|}{0.1484} & \multicolumn{1}{c|}{0.1671} & 0.1811 & \multicolumn{1}{c|}{0.0228} & \multicolumn{1}{c|}{0.0294} & \multicolumn{1}{c|}{0.0385} & \multicolumn{1}{c|}{0.0551} & \multicolumn{1}{c|}{0.0635} & 0.0703  \\ \hline
Collaborative filtering & \multicolumn{1}{c|}{0.0503} & \multicolumn{1}{c|}{0.0535} & \multicolumn{1}{c|}{0.0650} & \multicolumn{1}{c|}{0.0896} & \multicolumn{1}{c|}{0.1037} & 0.1135 & \multicolumn{1}{c|}{0.0276} & \multicolumn{1}{c|}{0.0349} & \multicolumn{1}{c|}{0.0444} & \multicolumn{1}{c|}{0.0581} & \multicolumn{1}{c|}{0.0651} & 0.0706  \\ \hline
Content Base            & \multicolumn{1}{c|}{0.0887} & \multicolumn{1}{c|}{0.0996} & \multicolumn{1}{c|}{0.1209} & \multicolumn{1}{c|}{0.1592} & \multicolumn{1}{c|}{0.1765} & 0.1892 & \multicolumn{1}{c|}{0.0305} & \multicolumn{1}{c|}{0.0388} & \multicolumn{1}{c|}{0.0499} & \multicolumn{1}{c|}{0.0686} & \multicolumn{1}{c|}{0.0778} & 0.0848  \\ \hline
Llm                     & \multicolumn{1}{c|}{0.0922} & \multicolumn{1}{c|}{0.1037} & \multicolumn{1}{c|}{0.1261} & \multicolumn{1}{c|}{0.1620} & \multicolumn{1}{c|}{0.1791} & 0.1924 & \multicolumn{1}{c|}{0.0353} & \multicolumn{1}{c|}{0.0430} & \multicolumn{1}{c|}{0.0529} & \multicolumn{1}{c|}{0.0696} & \multicolumn{1}{c|}{0.0783} & 0.0845  \\ \hline
Sequential              & \multicolumn{1}{c|}{0.0926} & \multicolumn{1}{c|}{0.1038} & \multicolumn{1}{c|}{0.1237} & \multicolumn{1}{c|}{0.1608} & \multicolumn{1}{c|}{0.1789} & 0.1920 & \multicolumn{1}{c|}{0.0341} & \multicolumn{1}{c|}{0.0423} & \multicolumn{1}{c|}{0.0527} & \multicolumn{1}{c|}{0.0697} & \multicolumn{1}{c|}{0.0782} & 0.0843  \\ \hline
Green                   & \multicolumn{1}{c|}{0.0922} & \multicolumn{1}{c|}{0.1037} & \multicolumn{1}{c|}{0.1261} & \multicolumn{1}{c|}{0.1620} & \multicolumn{1}{c|}{0.1791} & 0.1924 & \multicolumn{1}{c|}{0.0344} & \multicolumn{1}{c|}{0.0425} & \multicolumn{1}{c|}{0.0530} & \multicolumn{1}{c|}{0.0691} & \multicolumn{1}{c|}{0.0776} & 0.0843  \\ \hline
Multi modal             & \multicolumn{1}{c|}{0.0940} & \multicolumn{1}{c|}{0.1048} & \multicolumn{1}{c|}{0.1268} & \multicolumn{1}{c|}{0.1613} & \multicolumn{1}{c|}{0.1789} & 0.1912 & \multicolumn{1}{c|}{0.0356} & \multicolumn{1}{c|}{0.0434} & \multicolumn{1}{c|}{0.0541} & \multicolumn{1}{c|}{0.0700} & \multicolumn{1}{c|}{0.0784} & 0.0844  \\ \hline
fair                    & \multicolumn{1}{c|}{0.1080} & \multicolumn{1}{c|}{0.1190} & \multicolumn{1}{c|}{0.1390} & \multicolumn{1}{c|}{0.1736} & \multicolumn{1}{c|}{0.1916} & 0.2035 & \multicolumn{1}{c|}{0.0402} & \multicolumn{1}{c|}{\textbf{0.0484}} & \multicolumn{1}{c|}{\textbf{0.0590}} & \multicolumn{1}{c|}{0.0760} & \multicolumn{1}{c|}{0.0844} & 0.0908  \\ \hline
LSTM                    & \multicolumn{1}{c|}{0.0944} & \multicolumn{1}{c|}{0.1044} & \multicolumn{1}{c|}{0.1213} & \multicolumn{1}{c|}{0.1620} & \multicolumn{1}{c|}{0.1791} & 0.1924 & \multicolumn{1}{c|}{0.0273} & \multicolumn{1}{c|}{0.0345} & \multicolumn{1}{c|}{0.0442} & \multicolumn{1}{c|}{0.0583} & \multicolumn{1}{c|}{0.0652} & 0.0702  \\ \hline
GRU                     & \multicolumn{1}{c|}{0.0986} & \multicolumn{1}{c|}{0.1053} & \multicolumn{1}{c|}{0.1223} & \multicolumn{1}{c|}{0.1620} & \multicolumn{1}{c|}{0.1791} & 0.1924 & \multicolumn{1}{c|}{0.0305} & \multicolumn{1}{c|}{0.0384} & \multicolumn{1}{c|}{0.0494} & \multicolumn{1}{c|}{0.0681} & \multicolumn{1}{c|}{0.0774} & 0.0841  \\ \hline
IndRNN                  & \multicolumn{1}{c|}{0.0919} & \multicolumn{1}{c|}{0.1063} & \multicolumn{1}{c|}{0.1264} & \multicolumn{1}{c|}{0.1620} & \multicolumn{1}{c|}{0.1791} & 0.1924 & \multicolumn{1}{c|}{0.0341} & \multicolumn{1}{c|}{0.0420} & \multicolumn{1}{c|}{0.0535} & \multicolumn{1}{c|}{0.0692} & \multicolumn{1}{c|}{0.0771} & 0.0842  \\ \hline
\textbf{Proposed model}          & \multicolumn{1}{c|}{\textbf{0.1018}} & \multicolumn{1}{c|}{\textbf{0.1199}} & \multicolumn{1}{c|}{\textbf{0.1379}} & \multicolumn{1}{c|}{\textbf{0.1787}} & \multicolumn{1}{c|}{\textbf{0.1970}} & \textbf{0.2035} & \multicolumn{1}{c|}{\textbf{0.0406}} & \multicolumn{1}{c|}{0.0412} & \multicolumn{1}{c|}{0.0534} & \multicolumn{1}{c|}{\textbf{0.0786}} & \multicolumn{1}{c|}{\textbf{0.0855}} & \textbf{0.09094} \\ \hline
Dataset                 & \multicolumn{6}{c|}{MedicalRec III}                                                                                                                            & \multicolumn{6}{c|}{MedicalRec IV}                                                                                                                            \\ \hline
Measure\%               & \multicolumn{1}{c|}{D@5}    & \multicolumn{1}{c|}{D@10}   & \multicolumn{1}{c|}{D@20}   & \multicolumn{1}{c|}{D@50}   & \multicolumn{1}{c|}{D@75}   & D@100  & \multicolumn{1}{c|}{D@5}    & \multicolumn{1}{c|}{D@10}   & \multicolumn{1}{c|}{D@20}   & \multicolumn{1}{c|}{D@50}   & \multicolumn{1}{c|}{D@75}   & D@100   \\ \hline
Hybrid                  & \multicolumn{1}{c|}{0.1080} & \multicolumn{1}{c|}{0.1180} & \multicolumn{1}{c|}{0.1379} & \multicolumn{1}{c|}{0.1724} & \multicolumn{1}{c|}{0.1906} & 0.2032 & \multicolumn{1}{c|}{0.0402} & \multicolumn{1}{c|}{0.0480} & \multicolumn{1}{c|}{0.0583} & \multicolumn{1}{c|}{0.0753} & \multicolumn{1}{c|}{0.0839} & 0.0900  \\ \hline
SVD                     & \multicolumn{1}{c|}{0.1077} & \multicolumn{1}{c|}{0.1190} & \multicolumn{1}{c|}{0.1377} & \multicolumn{1}{c|}{0.1734} & \multicolumn{1}{c|}{0.1909} & 0.2028 & \multicolumn{1}{c|}{0.0400} & \multicolumn{1}{c|}{0.0484} & \multicolumn{1}{c|}{0.0583} & \multicolumn{1}{c|}{0.0755} & \multicolumn{1}{c|}{0.0839} & 0.0901  \\ \hline
Collaborative filtering & \multicolumn{1}{c|}{0.1076} & \multicolumn{1}{c|}{0.1188} & \multicolumn{1}{c|}{0.1390} & \multicolumn{1}{c|}{0.1733} & \multicolumn{1}{c|}{0.1909} & 0.2029 & \multicolumn{1}{c|}{0.0400} & \multicolumn{1}{c|}{0.0483} & \multicolumn{1}{c|}{0.0590} & \multicolumn{1}{c|}{0.0759} & \multicolumn{1}{c|}{0.0837} & 0.0900  \\ \hline
Content Base            & \multicolumn{1}{c|}{0.1062} & \multicolumn{1}{c|}{0.1167} & \multicolumn{1}{c|}{0.1364} & \multicolumn{1}{c|}{0.1736} & \multicolumn{1}{c|}{0.1901} & 0.2020 & \multicolumn{1}{c|}{0.0398} & \multicolumn{1}{c|}{0.0481} & \multicolumn{1}{c|}{0.0587} & \multicolumn{1}{c|}{0.0760} & \multicolumn{1}{c|}{0.0842} & 0.0907  \\ \hline
Llm                     & \multicolumn{1}{c|}{0.1073} & \multicolumn{1}{c|}{0.1179} & \multicolumn{1}{c|}{0.1387} & \multicolumn{1}{c|}{0.1734} & \multicolumn{1}{c|}{0.1916} & 0.2031 & \multicolumn{1}{c|}{0.0397} & \multicolumn{1}{c|}{0.0481} & \multicolumn{1}{c|}{0.0587} & \multicolumn{1}{c|}{0.0759} & \multicolumn{1}{c|}{0.0844} & 0.0907  \\ \hline
Sequential              & \multicolumn{1}{c|}{0.1071} & \multicolumn{1}{c|}{0.1177} & \multicolumn{1}{c|}{0.1375} & \multicolumn{1}{c|}{0.1727} & \multicolumn{1}{c|}{0.1904} & 0.2035 & \multicolumn{1}{c|}{0.0399} & \multicolumn{1}{c|}{0.0481} & \multicolumn{1}{c|}{0.0587} & \multicolumn{1}{c|}{0.0759} & \multicolumn{1}{c|}{0.0843} & 0.0908  \\ \hline
Green                   & \multicolumn{1}{c|}{0.1012} & \multicolumn{1}{c|}{0.1183} & \multicolumn{1}{c|}{0.1376} & \multicolumn{1}{c|}{0.1726} & \multicolumn{1}{c|}{0.1902} & 0.2034 & \multicolumn{1}{c|}{0.0403} & \multicolumn{1}{c|}{0.0483} & \multicolumn{1}{c|}{0.0585} & \multicolumn{1}{c|}{0.0752} & \multicolumn{1}{c|}{0.0831} & 0.0901  \\ \hline
Multi modal             & \multicolumn{1}{c|}{0.1072} & \multicolumn{1}{c|}{0.1191} & \multicolumn{1}{c|}{0.1375} & \multicolumn{1}{c|}{0.1734} & \multicolumn{1}{c|}{0.1903} & 0.2026 & \multicolumn{1}{c|}{0.0401} & \multicolumn{1}{c|}{0.0474} & \multicolumn{1}{c|}{0.0585} & \multicolumn{1}{c|}{0.0753} & \multicolumn{1}{c|}{0.0832} & 0.0903  \\ \hline
fair                    & \multicolumn{1}{c|}{0.1076} & \multicolumn{1}{c|}{0.1118} & \multicolumn{1}{c|}{0.1396} & \multicolumn{1}{c|}{0.1735} & \multicolumn{1}{c|}{0.1905} & 0.2027 & \multicolumn{1}{c|}{0.0405} & \multicolumn{1}{c|}{0.0475} & \multicolumn{1}{c|}{0.0591} & \multicolumn{1}{c|}{0.0754} & \multicolumn{1}{c|}{0.0833} & 0.0912  \\ \hline
LSTM                    & \multicolumn{1}{c|}{0.1083} & \multicolumn{1}{c|}{0.1181} & \multicolumn{1}{c|}{0.1374} & \multicolumn{1}{c|}{0.1723} & \multicolumn{1}{c|}{0.1906} & 0.2038 & \multicolumn{1}{c|}{0.0401} & \multicolumn{1}{c|}{0.0482} & \multicolumn{1}{c|}{0.0583} & \multicolumn{1}{c|}{0.0758} & \multicolumn{1}{c|}{0.0833} & 0.0914  \\ \hline
GRU                     & \multicolumn{1}{c|}{0.1073} & \multicolumn{1}{c|}{0.1191} & \multicolumn{1}{c|}{0.1373} & \multicolumn{1}{c|}{0.1734} & \multicolumn{1}{c|}{0.1906} & 0.2020 & \multicolumn{1}{c|}{0.0408} & \multicolumn{1}{c|}{0.0480} & \multicolumn{1}{c|}{0.0585} & \multicolumn{1}{c|}{0.0750} & \multicolumn{1}{c|}{0.0839} & 0.0911  \\ \hline
IndRNN                  & \multicolumn{1}{c|}{0.1071} & \multicolumn{1}{c|}{0.1182} & \multicolumn{1}{c|}{0.1395} & \multicolumn{1}{c|}{0.1734} & \multicolumn{1}{c|}{0.1909} & 0.2021 & \multicolumn{1}{c|}{0.0491} & \multicolumn{1}{c|}{0.0491} & \multicolumn{1}{c|}{0.0592} & \multicolumn{1}{c|}{0.0757} & \multicolumn{1}{c|}{0.0837} & 0.09190 \\ \hline
\textbf{Proposed model }         & \multicolumn{1}{c|}{\textbf{0.1101}} & \multicolumn{1}{c|}{\textbf{0.1235}} & \multicolumn{1}{c|}{\textbf{0.1404}} & \multicolumn{1}{c|}{\textbf{0.1798}} & \multicolumn{1}{c|}{\textbf{0.2001}} & \textbf{0.2101} & \multicolumn{1}{c|}{\textbf{0.0512}} & \multicolumn{1}{c|}{\textbf{0.0511}} & \multicolumn{1}{c|}{\textbf{0.0687}} & \multicolumn{1}{c|}{\textbf{0.0867}} & \multicolumn{1}{c|}{\textbf{0.0901}} & \textbf{0.09200} \\ \hline
\end{tabular}
\label{Table:2R}
\end{table*}

\begin{table*}[]
\centering
\caption{Results obtained by the proposed approach and baseline approaches on the MedicalRec I, MedicalRec II, MedicalRec III and MedicalRec IV datasets. Results are reported for the HitRate@100, Recall@100, Precision@100, and F1@100.}
\begin{tabular}{|l|cccc|cccc|}
\hline
Dataset                 & \multicolumn{4}{c|}{MedicalRec I}                                                                                   & \multicolumn{4}{c|}{MedicalRec II}                                                                                  \\ \hline
Measure\%               & \multicolumn{1}{c|}{HitRate@100} & \multicolumn{1}{c|}{Recall @100} & \multicolumn{1}{c|}{Precision @100} & F1 @100 & \multicolumn{1}{c|}{HitRate@100} & \multicolumn{1}{c|}{Recall @100} & \multicolumn{1}{c|}{Precision @100} & F1 @100 \\ \hline
Hybrid                  & \multicolumn{1}{c|}{60.27\%}     & \multicolumn{1}{c|}{3.24\%}      & \multicolumn{1}{c|}{2.00\%}         & 2.43\%  & \multicolumn{1}{c|}{67.50\%}     & \multicolumn{1}{c|}{9.08\%}      & \multicolumn{1}{c|}{9.41\%}         & 8.99\%  \\ \hline
SVD                     & \multicolumn{1}{c|}{60.88\%}     & \multicolumn{1}{c|}{2.85\%}      & \multicolumn{1}{c|}{1.83\%}         & 2.18\%  & \multicolumn{1}{c|}{68.43\%}     & \multicolumn{1}{c|}{8.93\%}      & \multicolumn{1}{c|}{9.65\%}         & 8.98\%  \\ \hline
Collaborative filtering & \multicolumn{1}{c|}{65.60\%}     & \multicolumn{1}{c|}{3.66\%}      & \multicolumn{1}{c|}{2.30\%}         & 2.77\%  & \multicolumn{1}{c|}{69.44\%}     & \multicolumn{1}{c|}{9.33\%}      & \multicolumn{1}{c|}{9.83\%}         & 9.29\%  \\ \hline
Content Base            & \multicolumn{1}{c|}{66.97\%}     & \multicolumn{1}{c|}{3.57\%}      & \multicolumn{1}{c|}{2.25\%}         & 2.70\%  & \multicolumn{1}{c|}{70.33\%}     & \multicolumn{1}{c|}{9.07\%}      & \multicolumn{1}{c|}{9.58\%}         & 9.04\%  \\ \hline
Llm                     & \multicolumn{1}{c|}{67.30\%}     & \multicolumn{1}{c|}{3.71\%}      & \multicolumn{1}{c|}{2.33\%}         & 2.80\%  & \multicolumn{1}{c|}{70.54\%}     & \multicolumn{1}{c|}{9.42\%}      & \multicolumn{1}{c|}{10.02\%}        & 9.41\%  \\ \hline
Sequential              & \multicolumn{1}{c|}{67.55\%}     & \multicolumn{1}{c|}{3.49\%}      & \multicolumn{1}{c|}{2.20\%}         & 2.64\%  & \multicolumn{1}{c|}{72.19\%}     & \multicolumn{1}{c|}{9.38\%}      & \multicolumn{1}{c|}{9.93\%}         & 9.36\%  \\ \hline
Green                   & \multicolumn{1}{c|}{70.00\%}     & \multicolumn{1}{c|}{3.82\%}      & \multicolumn{1}{c|}{2.40\%}         & 2.88\%  & \multicolumn{1}{c|}{70.64\%}     & \multicolumn{1}{c|}{3.83\%}      & \multicolumn{1}{c|}{2.41\%}         & 2.89\%  \\ \hline
Multi modal             & \multicolumn{1}{c|}{70.72\%}     & \multicolumn{1}{c|}{3.86\%}      & \multicolumn{1}{c|}{2.44\%}         & 2.93\%  & \multicolumn{1}{c|}{70.21\%}     & \multicolumn{1}{c|}{3.77\%}      & \multicolumn{1}{c|}{2.37\%}         & 2.85\%  \\ \hline
fair                    & \multicolumn{1}{c|}{67.09\%}     & \multicolumn{1}{c|}{3.42\%}      & \multicolumn{1}{c|}{2.16\%}         & 2.59\%  & \multicolumn{1}{c|}{69.74\%}     & \multicolumn{1}{c|}{3.70\%}      & \multicolumn{1}{c|}{2.34\%}         & 2.80\%  \\ \hline
LSTM                    & \multicolumn{1}{c|}{70.24\%}     & \multicolumn{1}{c|}{3.75\%}      & \multicolumn{1}{c|}{2.37\%}         & 2.84\%  & \multicolumn{1}{c|}{70.72\%}     & \multicolumn{1}{c|}{3.86\%}      & \multicolumn{1}{c|}{2.44\%}         & 2.93\%  \\ \hline
GRU                     & \multicolumn{1}{c|}{70.30\%}     & \multicolumn{1}{c|}{3.75\%}        & \multicolumn{1}{c|}{2.33\%}           & 2.85    & \multicolumn{1}{c|}{70.83\%}       & \multicolumn{1}{c|}{8.83\%}        & \multicolumn{1}{c|}{8.92\%}           & 9.12\%    \\ \hline
IndRNN                  & \multicolumn{1}{c|}{70.54\%}       & \multicolumn{1}{c|}{3.77\%}        & \multicolumn{1}{c|}{2.38\%}           & 2.78\%    & \multicolumn{1}{c|}{70.90\%}       & \multicolumn{1}{c|}{8.93\%}        & \multicolumn{1}{c|}{8.76\%}           & 9.42\%    \\ \hline
\textbf{Proposed model  }        & \multicolumn{1}{c|}{\textbf{71.51\%}}       & \multicolumn{1}{c|}{\textbf{3.95\%}}        & \multicolumn{1}{c|}{\textbf{2.56\%}}           & \textbf{3.00\%}    & \multicolumn{1}{c|}{\textbf{72.51\%}}       & \multicolumn{1}{c|}{\textbf{9.60\%}}        & \multicolumn{1}{c|}{\textbf{10.23\%}}          & \textbf{9.60}    \\ \hline
Dataset                 & \multicolumn{4}{c|}{MedicalRec III}                                                                                 & \multicolumn{4}{c|}{MedicalRec IV}                                                                                  \\ \hline
Measure\%               & \multicolumn{1}{c|}{HitRate@100} & \multicolumn{1}{c|}{Recall @100} & \multicolumn{1}{c|}{Precision @100} & F1 @100 & \multicolumn{1}{c|}{HitRate@100} & \multicolumn{1}{c|}{Recall @100} & \multicolumn{1}{c|}{Precision @100} & F1 @100 \\ \hline
Hybrid                  & \multicolumn{1}{c|}{68.24\%}     & \multicolumn{1}{c|}{3.68\%}      & \multicolumn{1}{c|}{2.32\%}         & 2.79\%  & \multicolumn{1}{c|}{70.41\%}     & \multicolumn{1}{c|}{9.21\%}      & \multicolumn{1}{c|}{9.72\%}         & 9.18\%  \\ \hline
SVD                     & \multicolumn{1}{c|}{69.43\%}     & \multicolumn{1}{c|}{3.75\%}      & \multicolumn{1}{c|}{2.37\%}         & 2.84\%  & \multicolumn{1}{c|}{71.21\%}     & \multicolumn{1}{c|}{9.21\%}      & \multicolumn{1}{c|}{9.78\%}         & 9.20\%  \\ \hline
Collaborative filtering & \multicolumn{1}{c|}{70.72\%}     & \multicolumn{1}{c|}{3.86\%}      & \multicolumn{1}{c|}{2.44\%}         & 2.93\%  & \multicolumn{1}{c|}{73.25\%}     & \multicolumn{1}{c|}{9.47\%}      & \multicolumn{1}{c|}{10.13\%}        & 9.48\%  \\ \hline
Content Base            & \multicolumn{1}{c|}{73.13\%}     & \multicolumn{1}{c|}{3.83\%}      & \multicolumn{1}{c|}{2.45\%}         & 2.92\%  & \multicolumn{1}{c|}{74.33\%}     & \multicolumn{1}{c|}{9.68\%}      & \multicolumn{1}{c|}{10.42\%}        & 9.72\%  \\ \hline
Llm                     & \multicolumn{1}{c|}{65.41\%}     & \multicolumn{1}{c|}{3.38\%}      & \multicolumn{1}{c|}{2.14\%}         & 2.56\%  & \multicolumn{1}{c|}{70.07\%}     & \multicolumn{1}{c|}{9.05\%}      & \multicolumn{1}{c|}{9.60\%}         & 9.03\%  \\ \hline
Sequential              & \multicolumn{1}{c|}{70.50\%}     & \multicolumn{1}{c|}{3.80\%}      & \multicolumn{1}{c|}{2.40\%}         & 2.88\%  & \multicolumn{1}{c|}{73.00\%}     & \multicolumn{1}{c|}{9.40\%}      & \multicolumn{1}{c|}{10.00\%}        & 9.40\%  \\ \hline
Green                   & \multicolumn{1}{c|}{71.00\%}     & \multicolumn{1}{c|}{3.85\%}      & \multicolumn{1}{c|}{2.42\%}         & 2.90\%  & \multicolumn{1}{c|}{73.50\%}     & \multicolumn{1}{c|}{9.45\%}      & \multicolumn{1}{c|}{10.05\%}        & 9.45\%  \\ \hline
Multi modal             & \multicolumn{1}{c|}{72.00\%}     & \multicolumn{1}{c|}{3.90\%}      & \multicolumn{1}{c|}{2.46\%}         & 2.95\%  & \multicolumn{1}{c|}{74.00\%}     & \multicolumn{1}{c|}{9.50\%}      & \multicolumn{1}{c|}{10.15\%}        & 9.50\%  \\ \hline
fair                    & \multicolumn{1}{c|}{69.50\%}     & \multicolumn{1}{c|}{3.70\%}      & \multicolumn{1}{c|}{2.35\%}         & 2.82\%  & \multicolumn{1}{c|}{72.50\%}     & \multicolumn{1}{c|}{9.35\%}      & \multicolumn{1}{c|}{9.95\%}         & 9.35\%  \\ \hline
LSTM                    & \multicolumn{1}{c|}{70.24\%}     & \multicolumn{1}{c|}{3.75\%}      & \multicolumn{1}{c|}{2.37\%}         & 2.84\%  & \multicolumn{1}{c|}{72.72\%}     & \multicolumn{1}{c|}{9.42\%}      & \multicolumn{1}{c|}{10.00\%}        & 9.42\%  \\ \hline
GRU                     & \multicolumn{1}{c|}{70.21\%}     & \multicolumn{1}{c|}{3.74\%}      & \multicolumn{1}{c|}{2.36\%}         & 2.82\%  & \multicolumn{1}{c|}{72.80\%}     & \multicolumn{1}{c|}{9.44\%}      & \multicolumn{1}{c|}{10.01\%}        & 9.45\%  \\ \hline
IndRNN                  & \multicolumn{1}{c|}{70.89\%}     & \multicolumn{1}{c|}{3.92\%}      & \multicolumn{1}{c|}{2.64\%}         & 2.97\%  & \multicolumn{1}{c|}{73.03\%}     & \multicolumn{1}{c|}{9.76\%}      & \multicolumn{1}{c|}{10.45\%}        & 9.94\%  \\ \hline
\textbf{Proposed model}          & \multicolumn{1}{c|}{\textbf{74.50\%} }    & \multicolumn{1}{c|}{\textbf{4.02\%}}      & \multicolumn{1}{c|}{\textbf{2.78\%}}         & \textbf{3.00\%}  & \multicolumn{1}{c|}{\textbf{75.50\%}}     & \multicolumn{1}{c|}{\textbf{9.80\%} }     & \multicolumn{1}{c|}{\textbf{10.50\%}}        & \textbf{9.80\% } \\ \hline
\end{tabular}
\label{Table:3R}
\end{table*}

Table \ref{Table:3R} compares the performance of different recommender system models on four datasets: MedicalRec I, MedicalRec II, MedicalRec III, and MedicalRec IV. It examines the evaluation metrics (HitRate@100, Recall@100, Precision@100, and F1@100) and the performance of the models. A brief and comprehensive analysis is provided below: 
\begin{enumerate}
    \item \textbf{HitRate@100:} Indicates the percentage of cases where at least one recommended item in the list of 100 was correct. This metric indicates the model's ability to provide at least one relevant recommendation. 
    \item \textbf{Recall@100:} The percentage of relevant items that are present in the list of 100 recommended by the model. This metric focuses on the coverage of relevant items. 
    \item \textbf{Precision@100:} The percentage of recommended items in the list of 100 that are genuinely relevant. This metric indicates the accuracy of the recommendations.
    \item \textbf{F1@100:} Harmonic mean of Precision and Recall, which shows the balance between these two metrics. 
\end{enumerate}

\begin{figure*}
\centering
\subfloat[ MedicalRec I]{\includegraphics[width=0.5\textwidth]{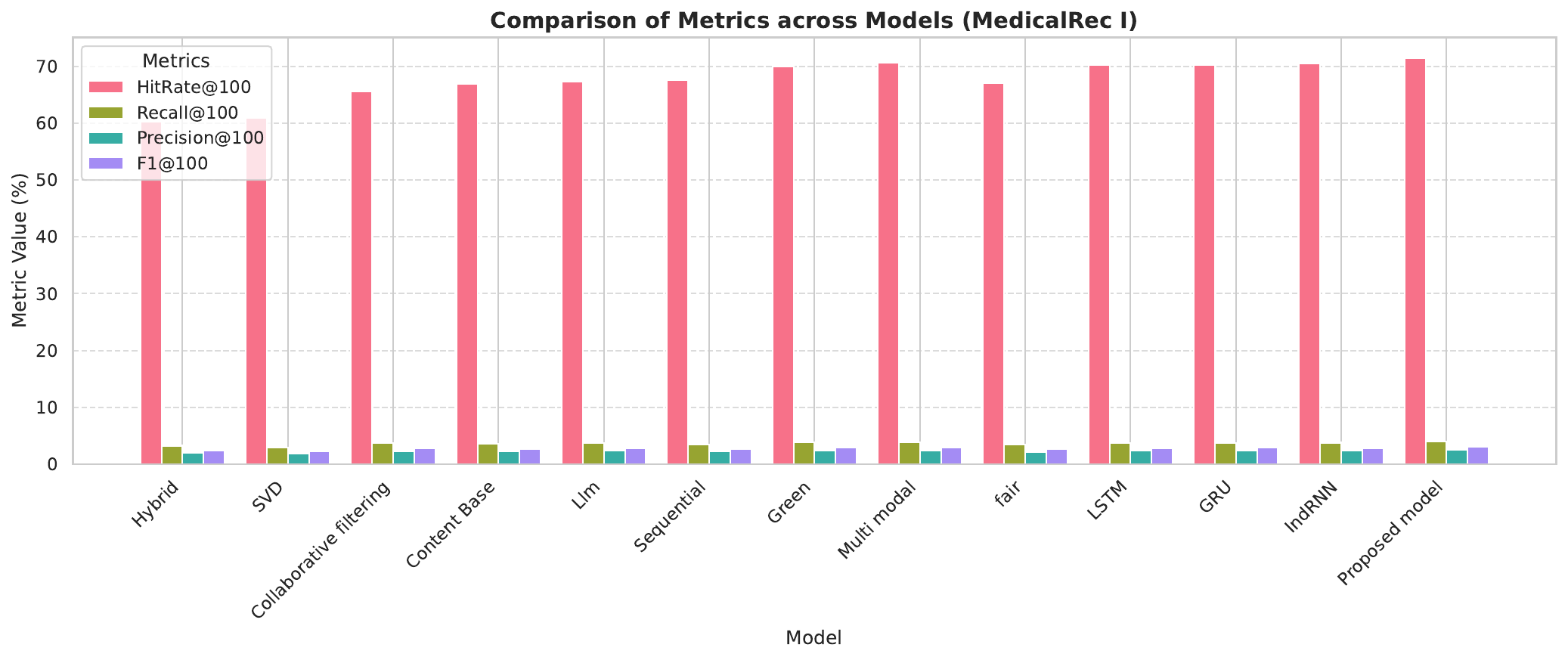}}
\subfloat[MedicalRec II]{\includegraphics[width=0.5\textwidth]{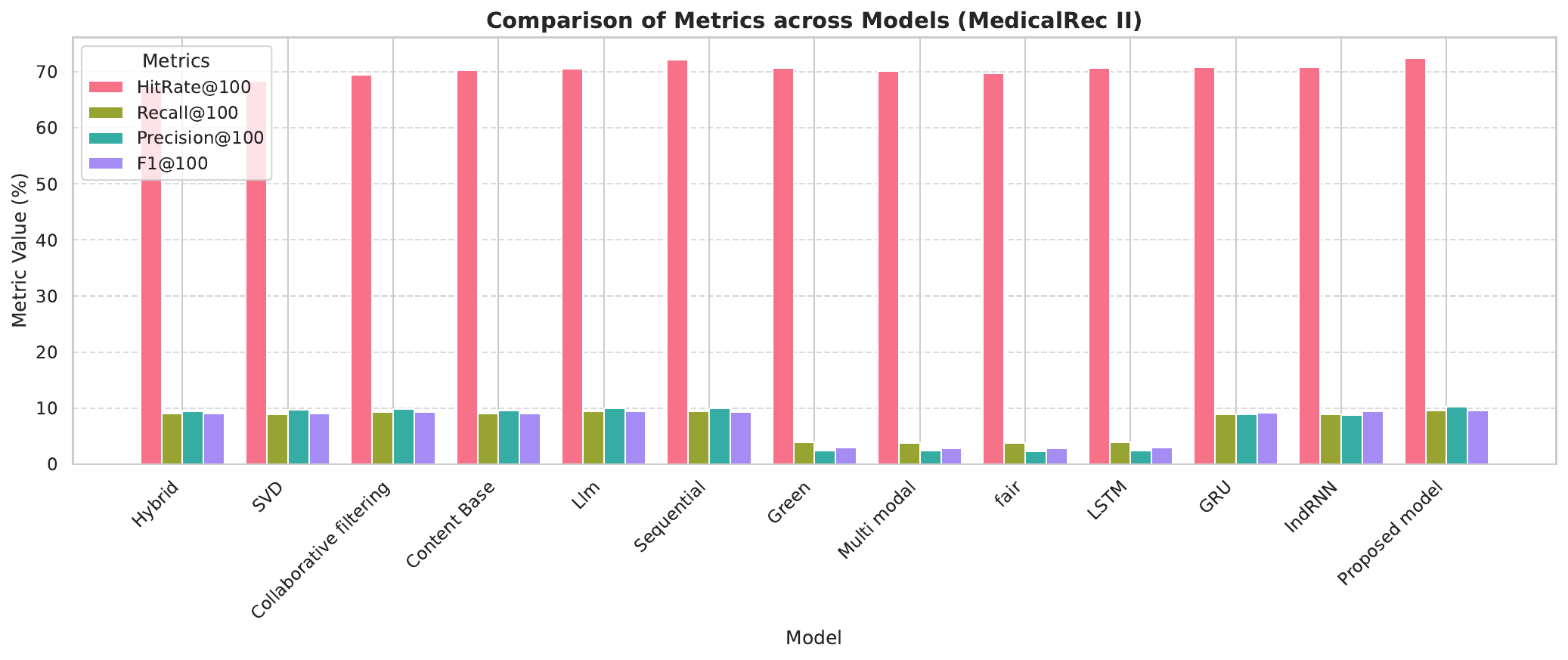}}\\
\subfloat[MedicalRec III]{\includegraphics[width=0.5\textwidth]{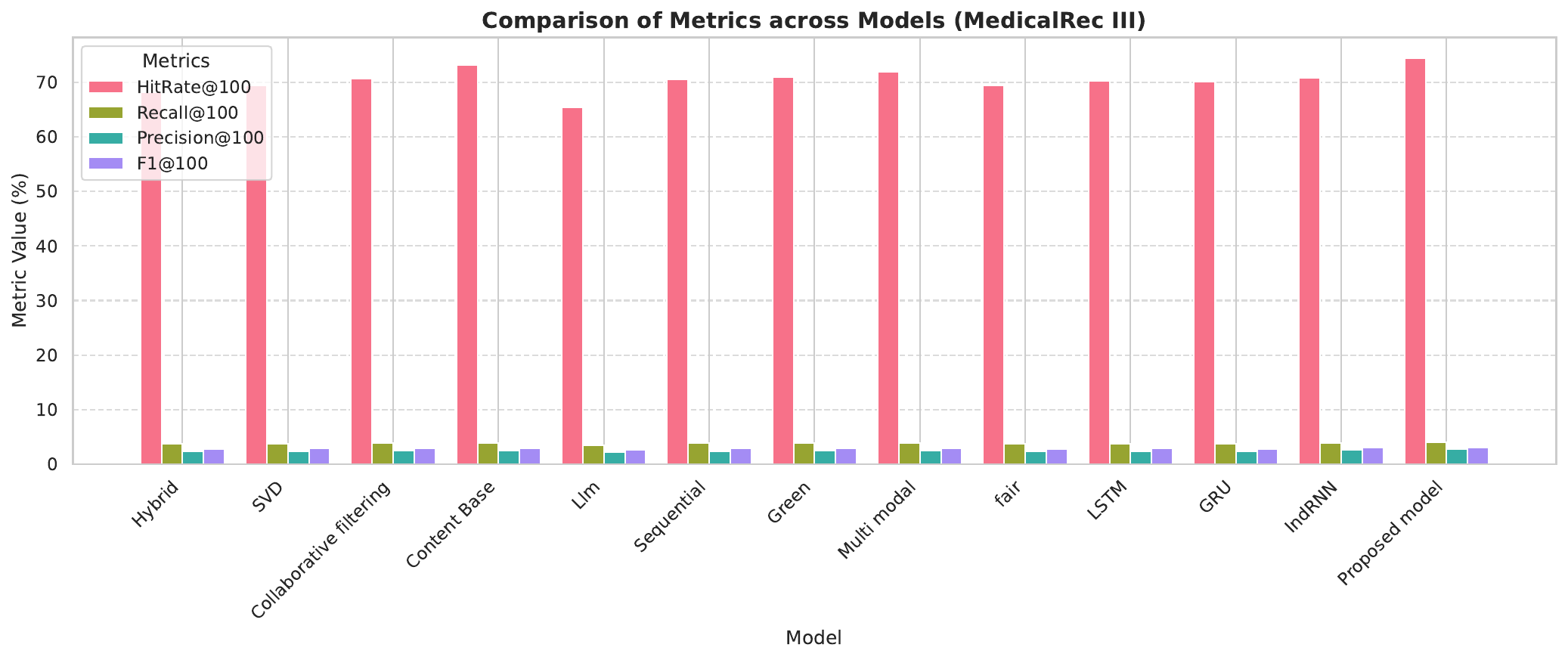}}
\subfloat[MedicalRec IV]{\includegraphics[width=0.5\textwidth]{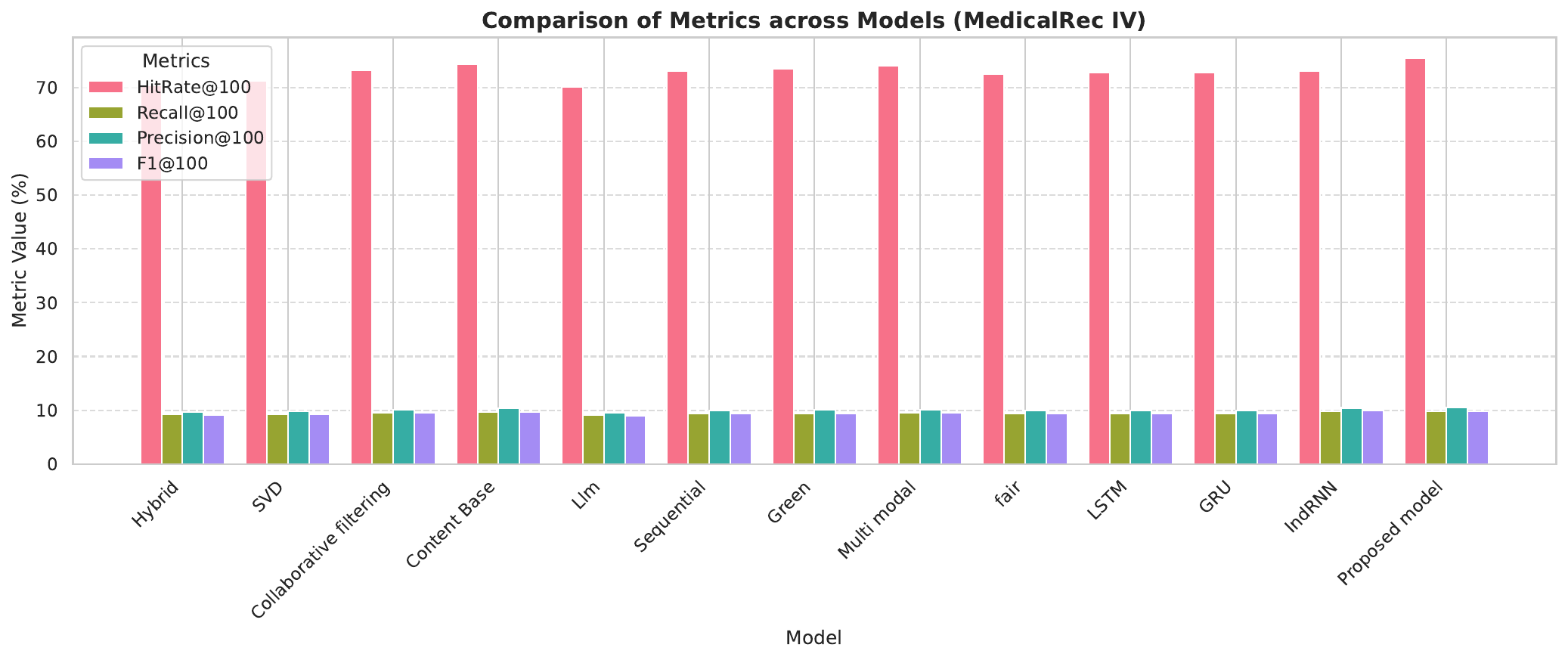}}\\
\caption{Comparison of Metrics across Models.}
\label{fig:fig7}
\end{figure*}

The proposed model performs best in all the metrics (HitRate, Recall, Precision, and F1) across all datasets (MedicalRec I to IV). The model achieved HitRate=71.51\%, Recall=3.95\%, Precision=2.56\%, F1=3.00\% in MedicalRec I, and HitRate=75.50\%, Recall=9.80\%, Precision=10.50\%, F1=9.80\% in MedicalRec IV. This indicates that the proposed model consistently performs well in identifying relevant and accurate items. The Hybrid and SVD models perform worse than the proposed model across all datasets. Especially in MedicalRec II, their Recall and Precision performances are significantly lower. Additionally, the two Collaborative Filtering and Content-Based models in MedicalRec III and IV outperform the Hybrid and SVD models, but are still weaker than the proposed model. The LLM approach in MedicalRec III performs worse than the other models (HitRate=65.41\%), but shows better performance in MedicalRec II and IV.
On the other hand, the Sequential, Green, Multi-modal, Fair, LSTM, GRU, IndRNN approaches perform close to each other in most criteria, but none of them reach the level of the proposed model. Especially in MedicalRec IV, the Multi-modal and IndRNN models are close to the proposed model, but still lag. In MedicalRec I, the overall performance of the models is lower (for example, Recall and Precision are about 2-4\%), which is mainly due to the reduction of the feature space. In MedicalRec II, we observe a significant improvement in Recall and Precision (up to approximately 9-10\%), indicating that this dataset is more suitable for recommender systems. Additionally, MedicalRec III and IV enhance the performance of the models, particularly in terms of HitRate and F1, underscoring the importance of utilizing a broader feature space. According to these results, it can be concluded that the proposed model performs best in all datasets and criteria, especially in MedicalRec IV, which has the highest values of HitRate (75.50\%), Recall (9.80\%), Precision (10.50\%), and F1 (9.80\%). On the other hand, MedicalRec IV seems to be the most suitable dataset for recommender systems, while MedicalRec I is more challenging. Also, the two Hybrid and SVD models have poorer performance and are not recommended for more complex scenarios. The comparative chart resulting from this analysis is shown in Figure \ref{fig:fig7}.

\subsection{K-fold cross validation}
For the k-fold analysis on the dataset, only the results for MedicalRec IV were reported. Table \ref{TBL:Kfold} presents the results obtained for the HitRate@100 criterion on this dataset. The proposed model consistently has the highest HitRate@100 in all folds (between 75.22\% and 76.50\%), with an average of approximately 75.67\%. It has a relative advantage over other models. The highest value obtained by this approach is in Fold-5. This model exhibits the least fluctuation between folds, indicating its good generalizability. The Content-Based model is ranked second and has a performance close to the proposed model. Among the tested models, the Hybrid and SVD approaches have the lowest performance, which may be due to the limitations of traditional methods or insufficient optimization in the Hybrid model. Among the Folds, Fold-5 has the highest average HitRate@100 among all models, which may be due to the distribution of the test data in this fold. On the other hand, Fold-1 has the lowest average HitRate@100, which may indicate that the test data in this fold is more challenging. Since the values of the Folds are chosen randomly, the results may vary in different runs.

\begin{table}[]
\centering
\caption{K-fold cross validation on MedicalRec IV.}
\tiny
\begin{tabular}{|l|l|l|l|l|l|}
\hline
Model                     & Fold-1  & Fold-2  & Fold-3  & Fold-4  & Fold-5  \\ \hline
Hybrid                    & 70.41\% & 71.23\% & 71.31\% & 70.78\% & 72.87\% \\ \hline
SVD                       & 71.32\% & 71.25\% & 71.21\% & 71.21\% & 73.21\% \\ \hline
Collaborative   filtering & 73.43\% & 73.42\% & 73.25\% & 73.25\% & 72.25\% \\ \hline
Content Base              & 74.23\% & 73.37\% & 74.33\% & 74.33\% & 74.33\% \\ \hline
Llm                       & 70.44\% & 72.67\% & 70.07\% & 70.07\% & 75.07\% \\ \hline
Sequential                & 73.05\% & 73.07\% & 73.00\% & 74.00\% & 74.00\% \\ \hline
Green                     & 73.56   & 73.56   & 73.50   & 74.53   & 75.56   \\ \hline
Multi modal               & 74.76\% & 73.04\% & 74.12\% & 74.03\% & 73.20\% \\ \hline
fair                      & 72.53\% & 73.53\% & 72.55\% & 72.52\% & 73.50\% \\ \hline
LSTM                      & 72.74\% & 73.42\% & 72.76\% & 73.72\% & 72.72\% \\ \hline
GRU                       & 72.53\% & 73.50\% & 72.82\% & 74.83\% & 72.80\% \\ \hline
IndRNN                    & 73.23\% & 74.53\% & 73.43\% & 73.44\% & 74.03\% \\ \hline
Proposed   model          & 75.22\% & 75.56\% & 75.53\% & 75.54\% & 76.50\% \\ \hline
\end{tabular}
\label{TBL:Kfold}
\end{table}

The results of the t-test (P-value) for each model are given in Table \ref{TBL2}. The 5-fold results in Table \ref{TBL:Kfold} were used to obtain these values. For this purpose, a vector for each model was considered, which includes the HitRate@100 values of each model in each fold. The main diagonal of the table has a p-value = 1. Here, the P-value indicates the probability of observing the results obtained under the null hypothesis. Usually, a significance level of 0.05 (or 0.01) is used as the threshold to determine statistical significance. If the P-value is less than 0.05, the null hypothesis (which usually indicates no difference between the two models) is rejected, and we conclude that there is no significant difference in performance between the two models. Also, diagonal values are equal to 1.00, which indicates that there is no difference between a model and itself. According to the results obtained, it can be concluded that the Proposed model has very low P-values (often 0.0000 or close to it) compared to all other models (last column and last row). This indicates that the performance of this model is statistically significantly different from all other models.

\begin{table*}[]
\caption{T-test(P-value) results  from different approaches in 5-fold cross validation on MedicalRec IV.}
\resizebox{\textwidth}{!}{%
\begin{tabular}{|l|c|c|c|c|c|c|c|c|c|c|c|c|c|}
\hline
Model                     & Hybrid & SVD    & Collaborative   filtering & Content   Base & Llm    & Sequential & Green  & Multi   modal & fair   & LSTM   & GRU    & IndRNN & Proposed   model \\ \hline
Hybrid                    & \textcolor[rgb]{0.00,0.07,1.00}{1.0000} & 0.5933 & 0.0053                    & 0.0003         & 0.7552 & 0.0024     & 0.0013 & 0.0014        & 0.0106 & 0.0058 & 0.0102 & 0.0011 & 0.0000           \\ \hline
SVD                       & 0.5933 & \textcolor[rgb]{0.00,0.07,1.00}{1.0000} & 0.0111                    & 0.0005         & 0.9824 & 0.0046     & 0.0022 & 0.0025        & 0.0235 & 0.0123 & 0.0200 & 0.0019 & 0.0000           \\ \hline
Collaborative   filtering & 0.0053 & 0.0111 & \textcolor[rgb]{0.00,0.07,1.00}{1.0000}                    & 0.0088         & 0.1851 & 0.3740     & 0.0569 & 0.1035        & 0.5690 & 0.8785 & 0.7182 & 0.0993 & 0.0000           \\ \hline
Content   Base            & 0.0003 & 0.0005 & 0.0088                    & \textcolor[rgb]{0.00,0.07,1.00}{1.0000}         & 0.0393 & 0.0502     & 0.9583 & 0.4571        & 0.0045 & 0.0059 & 0.1092 & 0.2327 & 0.0006           \\ \hline
Llm                       & 0.7552 & 0.9824 & 0.1851                    & 0.0393         & \textcolor[rgb]{0.00,0.07,1.00}{1.0000} & 0.1188     & 0.0475 & 0.0685        & 0.2462 & 0.1974 & 0.1637 & 0.0751 & 0.0040           \\ \hline
Sequential                & 0.0024 & 0.0046 & 0.3740                    & 0.0502         & 0.1188 & \textcolor[rgb]{0.00,0.07,1.00}{1.0000}     & 0.1626 & 0.3339        & 0.1773 & 0.2958 & 0.7955 & 0.3938 & 0.0001           \\ \hline
Green                     & 0.0013 & 0.0022 & 0.0569                    & 0.9583         & 0.0475 & 0.1626     & \textcolor[rgb]{0.00,0.07,1.00}{1.0000} & 0.5598        & 0.0321 & 0.0462 & 0.1821 & 0.4004 & 0.0103           \\ \hline
Multi   modal             & 0.0014 & 0.0025 & 0.1035                    & 0.4571         & 0.0685 & 0.3339     & 0.5598 & \textcolor[rgb]{0.00,0.07,1.00}{1.0000}        & 0.0528 & 0.0809 & 0.3369 & 0.7998 & 0.0014           \\ \hline
fair                      & 0.0106 & 0.0235 & 0.5690                    & 0.0045         & 0.2462 & 0.1773     & 0.0321 & 0.0528        & \textcolor[rgb]{0.00,0.07,1.00}{1.0000} & 0.6589 & 0.4631 & 0.0458 & 0.0000           \\ \hline
LSTM                      & 0.0058 & 0.0123 & 0.8785                    & 0.0059         & 0.1974 & 0.2958     & 0.0462 & 0.0809        & 0.6589 & \textcolor[rgb]{0.00,0.07,1.00}{1.0000} & 0.6430 & 0.0730 & 0.0000           \\ \hline
GRU                       & 0.0102 & 0.0200 & 0.7182                    & 0.1092         & 0.1637 & 0.7955     & 0.1821 & 0.3369        & 0.4631 & 0.6430 & \textcolor[rgb]{0.00,0.07,1.00}{1.0000} & 0.3957 & 0.0010           \\ \hline
IndRNN                    & 0.0011 & 0.0019 & 0.0993                    & 0.2327         & 0.0751 & 0.3938     & 0.4004 & 0.7998        & 0.0458 & 0.0730 & 0.3957 & \textcolor[rgb]{0.00,0.07,1.00}{1.0000} & 0.0003           \\ \hline
Proposed model            & 0.0000 & 0.0000 & 0.0000                    & 0.0006         & 0.0040 & 0.0001     & 0.0103 & 0.0014        & 0.0000 & 0.0000 & 0.0010 & 0.0003 & \textcolor[rgb]{0.00,0.07,1.00}{1.0000}           \\ \hline
\end{tabular}}
\label{TBL2}
\end{table*}

\section{Conclusion}
Choosing the right model for medical image classification is challenging due to the variety of algorithms and different accuracies. Machine learning and deep learning models have been highly recommended in the literature for this purpose. Deep learning models are costly and time-consuming due to the carbon footprint associated with training billions of parameters. Model and benchmark providers often need to train and test multiple models to identify the best model for their datasets. The lack of a suitable benchmark in this area has created challenges. In this paper, we first attempted to collect the MedicalRec-bench dataset, which comprises more than 5,000 records from 3,000 articles on medical image classification. The dataset has significant missing values due to the lack of values provided by the authors and model providers. Next, a model called MedicalRec was introduced, which aims to provide a classification model. MedicalRec is essentially a Transformer that utilizes Softmax in the final layer to calculate the probability of each item. Twelve base models were used for evaluation on the MedicalRec dataset, which was divided into four categories: MedicalRec I (5 features), MedicalRec II (9 features), MedicalRec III (11 features), and MedicalRec IV (18 features). This dataset is the first to serve this purpose in the literature, utilizing statistical features derived from common literature models to provide a model recommendation system. To evaluate the proposed model, five evaluation metrics, Precision@K, Recall@K, F1@K, Discounted Cumulative Gain (DCG), and HitRate@K were used. The p-value obtained from 5-fold cross-validation of different approaches indicates the superiority of the proposed approach on the presented dataset. 

The dataset also includes titles, abstracts, and keywords of the articles, which can be used in training an LLM-based or Multi-modal model in the future. Zero starting values, missing values, and class imbalance are challenges that can be addressed in the future. The dataset and proposed model are made publicly available for future research.

%

\begin{IEEEbiography}[{\includegraphics[width=1in,height=1.25in,clip,keepaspectratio]{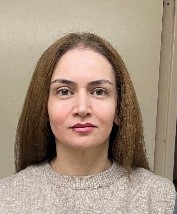}}]{Roghayeh Taghavi} received the Ph.D. degree in particle physics from the University of Yazd. She finished her first post-doc in 2019 at the University of Tehran. She is now finishing her second post-doc at Toronto Metropolitan University. She received a certificate in Data Analytics, Big Data, and Predictive Analytics in 2025 at TMU, Canada. She is also working as a machine learning engineer at Vosyn and as a senior AI trainer at Handshake (USA) remotely.
\end{IEEEbiography}

\begin{IEEEbiography}[{\includegraphics[width=1in,height=1.25in,clip,keepaspectratio]{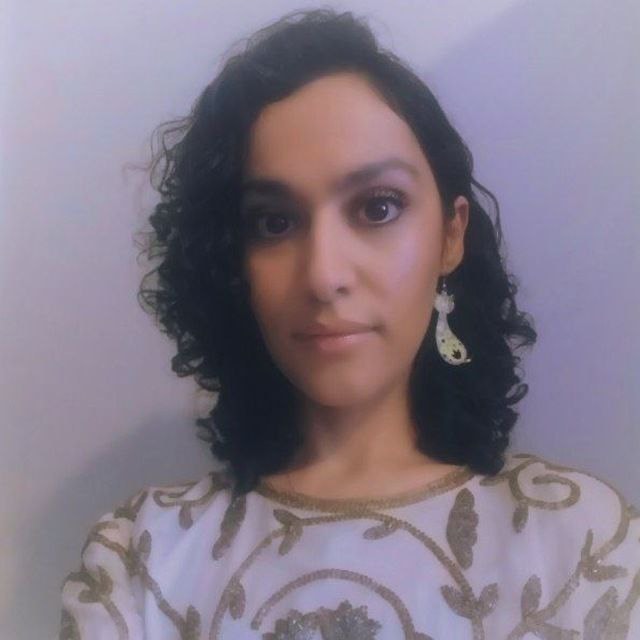}}]{Aysa Hasanazde Bashkandi} received the M.D. degree from Urmia University of Medical Sciences, Iran in 2014. She is currently an independent researcher specializing in medical imaging technologies. Since 2023, her scholarly contribution and research has been strategically directed towards the validation and implementation of AI-driven diagnostic algorithms, aiming to enhance precision and efficiency in radiological diagnostic processes. Her work represents a critical advancement in translational medical informatics, bridging computational methodologies with clinical imaging interpretation.
\end{IEEEbiography}

\begin{IEEEbiography}[{\includegraphics[width=1in,height=1.25in,clip,keepaspectratio]{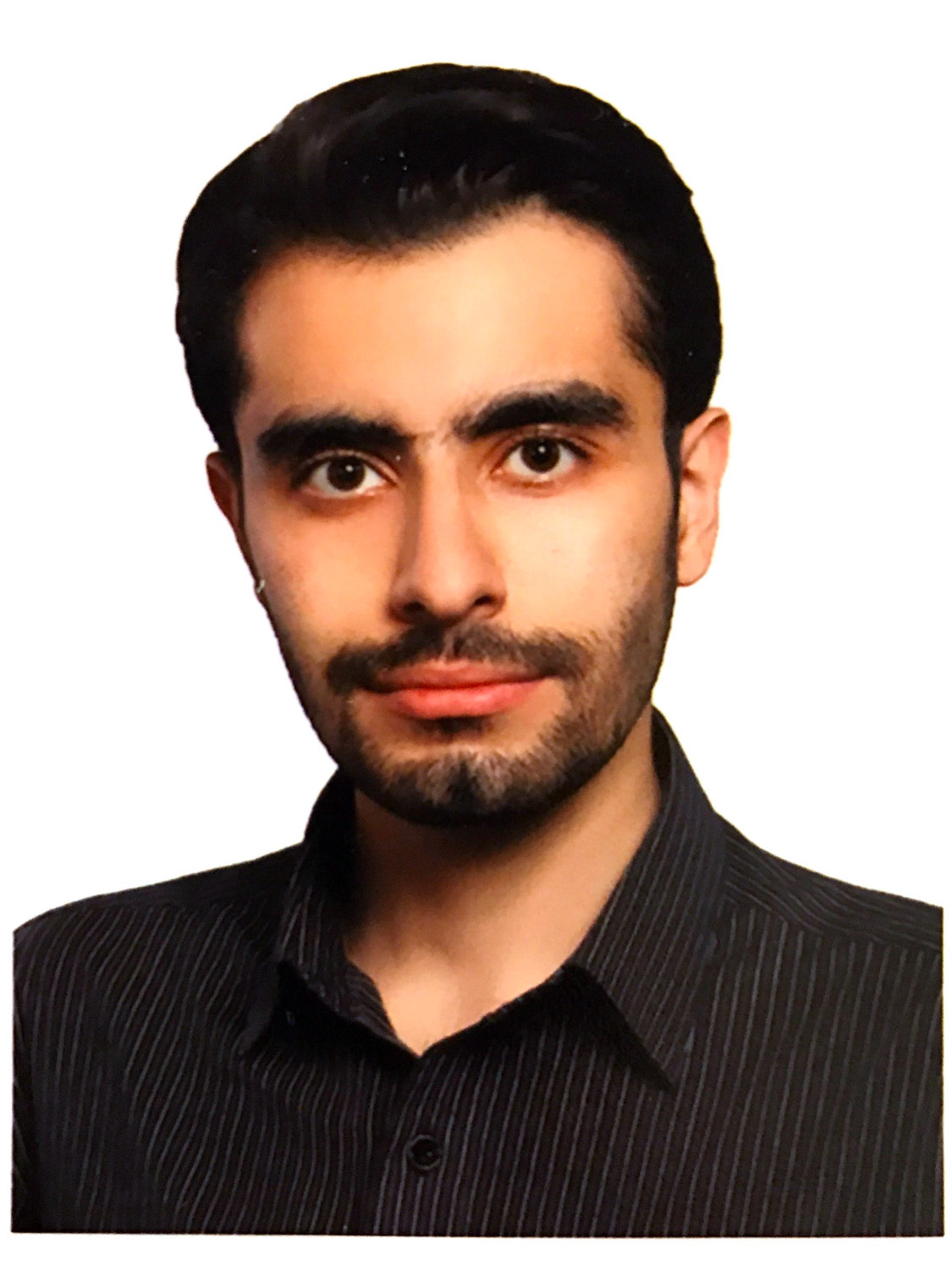}}]{Amir Ali Bengari}   received the B.Sc. degree in Civil Engineering from the University of Tehran, Tehran, Iran, in 2021, and the M.Sc. degree in Operations Research from the same university in 2024. He is currently a Research Assistant at the University of Tehran. His research interests include applications of deep learning in medicine and finance, multi‑agent reinforcement learning for power‑grid optimization, digital twins for infrastructure systems, and data‑driven optimization. He has been recognized for academic excellence during his studies.
\end{IEEEbiography}

\begin{IEEEbiography}[{\includegraphics[width=1in,height=1.25in,clip,keepaspectratio]{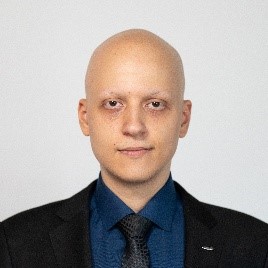}}]{Mohammad Amin Raji}  received his M.D. in medicine in 2023 from Babol University of Medical Sciences, Iran. His research interests include Artificial Intelligence, Computer Vision, Computational Neuroscience, and Brain-Computer Interfaces.
\end{IEEEbiography}
\begin{IEEEbiography}[{\includegraphics[width=1in,height=1.25in,clip,keepaspectratio]{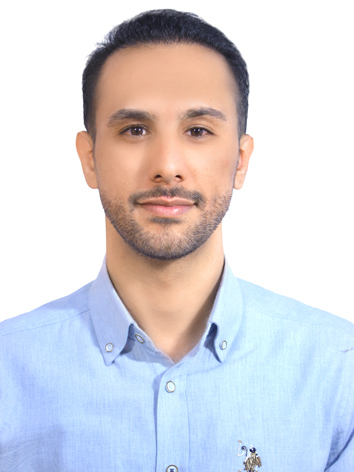}}]{Mohammad Salahi Ardekani}   received the B.Sc. degree in Software Engineering from Asharafi Isfahani University, Iran. He obtained his first M.Sc. degree in Software Engineering from Pooya Higher Education Institute, Iran, and is currently pursuing a second M.Sc. degree in International Business Management at OSTİM Teknik Üniversitesi, Ankara, Türkiye. His research interests include machine learning, intelligent software agents, predictive modeling, and optimization for decision-making systems. He has contributed to several international research and development projects integrating artificial intelligence with business analytics, and his current work focuses on leveraging AI-driven optimization for enhanced performance and predictive intelligence in complex systems.
\end{IEEEbiography}

\begin{IEEEbiography}[{\includegraphics[width=1in,height=1.25in,clip,keepaspectratio]{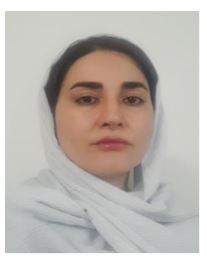}}]{Parisa Mardukhian} She received her Bachelor's degree in Software Engineering from University Jihad Institute of
Higher Education and continued her academic journey by obtaining a Master's degree in the
same field from Zagros University Bachelor of software engineering. Her primary research
interests lie at the intersection of Data Science and Artificial Intelligence, with a specific focus on
practical applications of Machine Learning. 
\end{IEEEbiography}
\begin{IEEEbiography}[{\includegraphics[width=1in,height=1.25in,clip,keepaspectratio]{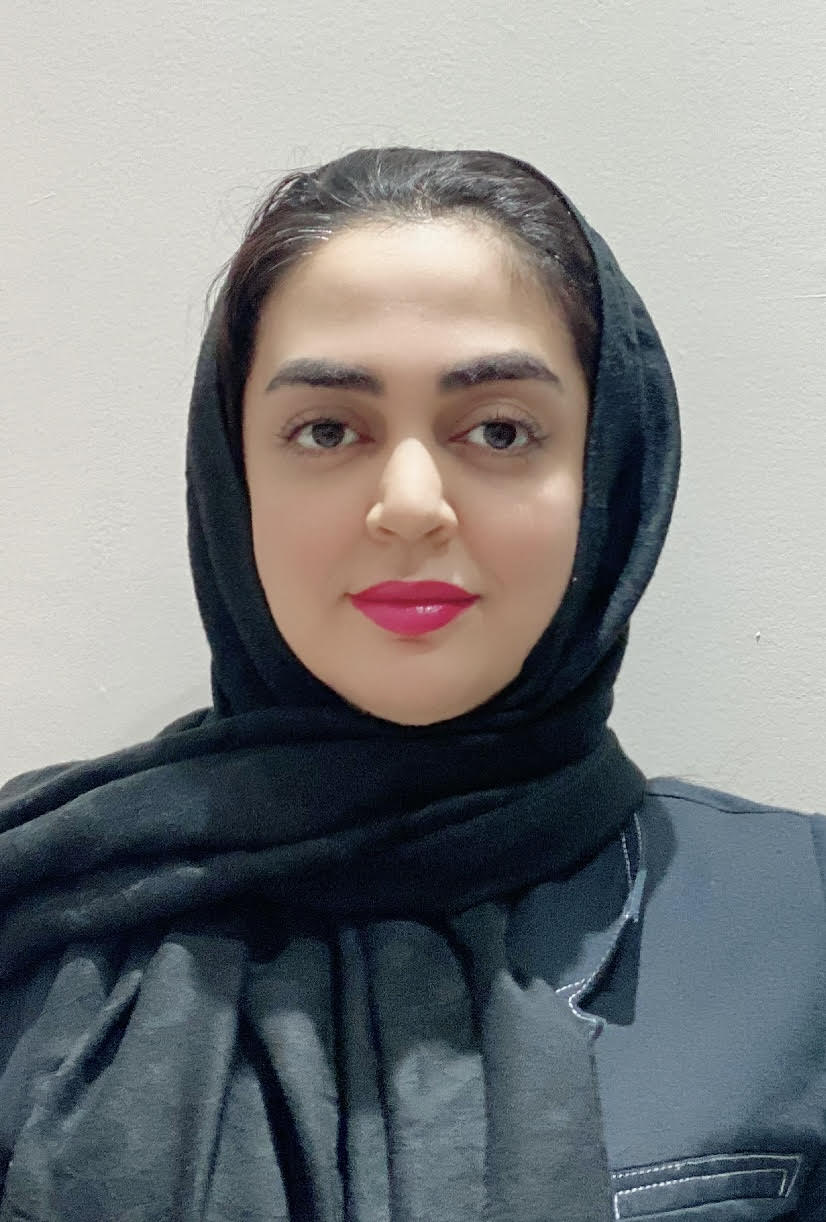}}]{Parvaneh Rezaei} received the B.Sc. degree in Information Technology from Payame‑Noor University and the M.Sc. degree in Artificial Intelligence from Salman University, Mashhad, Iran. She is currently a Medical Image Processing Specialist at Shana Surgery Clinic. Her research interests include image processing, face recognition, face embedding, generative adversarial networks (GANs), multi-task learning, and data science.
\end{IEEEbiography}

\begin{IEEEbiography}[{\includegraphics[width=1in,height=1.25in,clip,keepaspectratio]{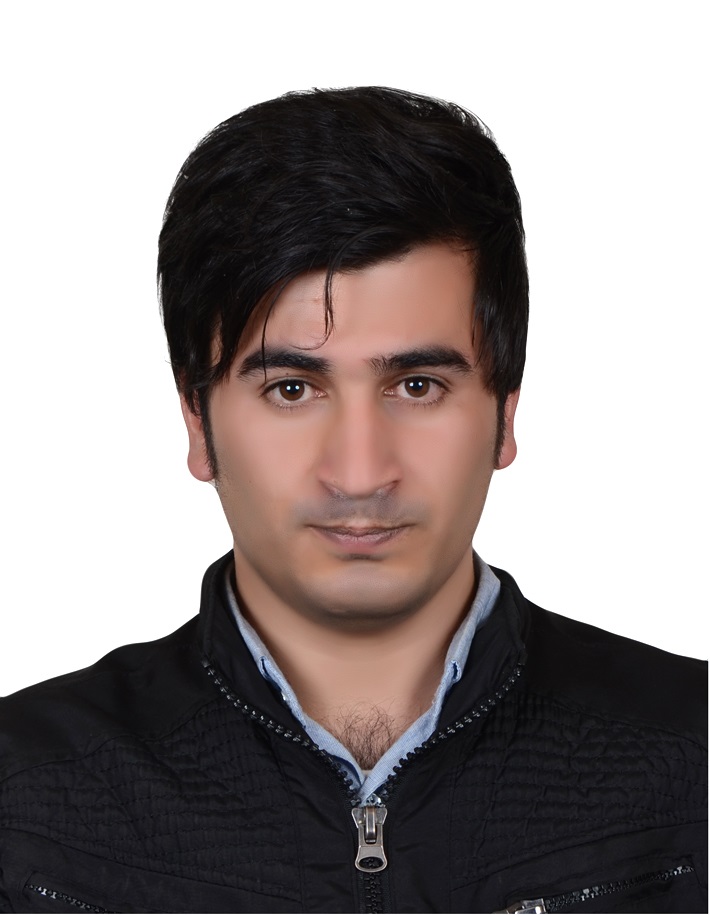}}]{Ramin Mousa}  received the master’s degree in Computer science from Zanjan university, Zanjan, Iran, in 2018. His research interests include topics in computer vision, machine learning, large language models
and medical image processing, 
\end{IEEEbiography}



\end{document}